%% file: main.tex
\documentclass[journal]{IEEEtran}

\usepackage[T1]{fontenc}
\usepackage{times}
\usepackage{dcolumn}
\usepackage{endnotes}
\usepackage{graphics}
\usepackage{times}
\usepackage{epsfig}
\usepackage{graphicx}
\usepackage{textcomp}
\usepackage{amssymb,amsmath}
\usepackage{balance}

\usepackage{listings}
\usepackage[]{algorithm2e}
\usepackage{flushend}
\usepackage{color}
\usepackage{hyperref}
\usepackage{verbatim} %comments
\usepackage{wrapfig}
\usepackage{siunitx}
\usepackage{todonotes}
\usepackage{eucal}
\usepackage{mathtools}
\usepackage[cal=dutchcal]{mathalfa}
\usepackage{bm}

\DeclareMathOperator*{\argmax}{arg\,max}
\DeclareMathOperator*{\argmin}{arg\,min}

\begin{document}

\title{
%Cellular Automaton for Efficient Randomization of Distributed Representations for Vector Symbolic Architectures
%Cellular Automata Enable Space-Time Tradeoff for Generation of Random Dense Binary Distributed Representations in Collective-State Computing
%Cellular Automata Enable Space-Time Tradeoff for Collective-State Computing
Cellular Automata Can Reduce Memory Requirements of Collective-State Computing %\\
%Cellular Automata Can Enable Space-Time Tradeoff for Collective-State Computing
}

\author{Denis Kleyko, E. Paxon Frady, and Friedrich T. Sommer
\thanks{
The authors thank members of the Redwood Center for Theoretical Neuroscience and Berkeley Wireless Research Center for stimulating discussions. 
The authors also thank Evgeny Osipov and Abbas Rahimi for fruitful discussions on the potential role of cellular automata in vector symbolic architectures, which inspired the current work.
The work of DK was supported by the European Union’s Horizon 2020 Research and Innovation Programme under the Marie Skłodowska-Curie Individual Fellowship Grant Agreement 839179 and in part by the DARPA’s VIP (Super-HD Project) and AIE (HyDDENN Project) programs.
FTS is supported by NIH R01-EB026955. 
} %revised January 9, 2020; accepted March 6, 2020. Date of publication April 06, 2020; date of current version March 6, 2020.}
%\thanks{Manuscript received December 26, 2016. This study  is  supported in
% part  by the Swedish Research Council (grant no. 2015-04677). DK also acknowledges Stiftelsen Seth M Kempes Stipendiefond for partially funding his research visit to UC Berkeley.}
\thanks{D. Kleyko is with the Redwood Center for Theoretical Neuroscience at the University of California, Berkeley, CA 94720, USA and also with the Intelligent Systems Lab at Research Institutes of Sweden, 164 40 Kista, Sweden. \mbox{E-mail}: \mbox{denkle@berkeley.edu}
}
\thanks{E. P. Frady and F. T. Sommer are with the Neuromorphic Computing Lab, Intel Labs and also with the Redwood Center for Theoretical Neuroscience at the University of California, Berkeley, CA 94720, USA. \mbox{E-mail}: \mbox{epaxon@berkeley.edu, fsommer@berkeley.edu}% 
}
}%

\markboth{}%
%\markboth{}%
{Kleyko \MakeLowercase{\textit{et al.}}: 
Cellular Automata Enable Space-Time Tradeoff}
%Cellular Automaton for Efficient Randomization}

\maketitle

\begin{abstract}
Various non-classical approaches of distributed information processing, such as neural networks, computation with Ising models, reservoir computing, vector symbolic architectures, and others, employ the principle of collective-state computing. In this type of computing, the variables relevant in a computation are superimposed into a single high-dimensional state vector, the collective-state. The variable encoding uses a fixed set of random patterns, which has to be stored and kept available during the computation.   
Here we show that an elementary cellular automaton with rule 90 (CA90) enables space-time tradeoff for collective-state computing models that use random dense binary representations, i.e., memory requirements can be traded off with computation running CA90.
We investigate the randomization behavior of CA90, in particular, the relation between the length of the randomization period and the size of the grid, and how CA90 preserves similarity in the presence of the initialization noise.
Based on these analyses we discuss how to optimize a collective-state computing model, in which
%Instead of memorizing the representations in their full length, the use of 
CA90 expands representations on the fly from short seed patterns -- 
rather than storing the full set of random patterns.
%while using only a fraction of the fully memorized solution at the cost of running CA90 computations every time when accessing the representations.
The CA90 expansion is applied and tested in concrete scenarios 
%several scenarios in the scope of 
using reservoir computing and vector symbolic architectures. 
Our experimental results show that collective-state computing with CA90 expansion performs similarly  
compared to traditional collective-state models, in which random patterns are generated initially by a pseudo-random number generator and then stored in a large memory. 
%results to storing the full patterns produced by 
%ded representations are functionally equivalent to the representations obtained from 
%a standard . 
\end{abstract}


\input{Introduction} 
\input{Concepts}

\input{CA90_Randomization}

\input{Results}

\input{Discussion}
%\input{tex/Conclusions}

\bibliography{references}

\appendices
\input{intESN}
\input{Resonator}

\end{document}

%% file: Introduction.tex
\section{Introduction}
\label{sect:intro}

Collective-state computing is an emerging paradigm of computing which leverages interactions of nodes or neurons in a highly interconnected network~\cite{CsabaCoupled2020}. This paradigm was first proposed in the context of neural networks and neuroscience for exploiting the parallelism of complex network dynamics to perform challenging computations. The classical examples include reservoir computing (RC)~\cite{LSM02, ESN02, RodanMinimumESN2011, FradyTheory2018} for buffering spatio-temporal inputs, and attractor networks for associative memory~\cite{hopfield1982neural} and optimization~\cite{hopfield1985neural}. 
In addition, many other models can be regarded as collective-state computing, such as random projection~\cite{AchlioptasDatabase2003}, compressed sensing~\cite{donoho2006, AminiDeterministic2011}, randomly connected feedforward neural networks~\cite{RVFLorig, KleykoDensityEncoding2020}, and vector symbolic architectures (VSAs)~\cite{PlateTr, Rachkovskij2001, Kanerva09}. Interestingly, these diverse computational models share a fundamental commonality -- they all include an initialization phase in which high-dimensional i.i.d. random vectors or matrices are generated that have to be memorized.
%, which then act as distributed representations and stay fixed 
%during the operation phase.
%, a kind of ``frozen randomness''. 
%We argue that the requirement of memorizing random vectors is a general hallmark of the paradigm of collective-state computing and serves the following purpose.
%that the collective states they use as basic elements of information representation random vectors, such as binary patterns with i.i.d. equal probabilities for ones and zeros.   
%Many of collective-state computing models use random dense binary distributed representations. Even though the aforementioned models are rarely considered all together, there is a deep relationship between them in the way they use randomness when forming the collective-state. 

In different models, these memorized random vectors serve a similar purpose: to represent
%At any instance of a computation, the collective state is a distributed representation of the different 
inputs and variables that need to be manipulated as distributed patterns across all neurons. The collective state is the (linear) superposition of these distributed representations. Decoding a particular variable from the collective state can be achieved by a linear projection onto the corresponding representation vector. Since high-dimensional random vectors are pseudo-orthogonal, the decoding interference is rather small, even if the collective state contains many variables\footnote{Although decoding by projection would work even better for exactly orthogonal distributed patterns, such a coding scheme is less desirable: i.i.d random generation of distributed patterns is computationally much cheaper and does not pose a hard limit on the number of encoded variables to be smaller or equal than the number of nodes or neurons.}. In contrast, if representations of different variables are not random, but contain correlations or statistical dependencies, then the interference becomes large when decoding and collective-state computing ceases to work. In order to achieve near orthogonality and low decoding interference, a large dimension of the random vectors is essential. 
%is constructed by superimposing different inputs to the system, which are first projected onto their corresponding distributed representations (``frozen'' randomness). This means that if one goes away from randomness the collective state stops working as it relies on the superposition of random representations. Therefore, during the whole system lifetime we have to store the ``frozen'' randomness generated at the initialization phase. 

%It seems strangely counter intuitive to spend a large amount of memory just to store random vectors. 
When implementing a collective-state computing system in hardware (e.g., in Field Programmable Gate Arrays, FPGA), the memory requirements are typically %required storage of large arrays of random vectors 
%one has to store random patterns, in order to enable the access of individual variables from the collective state. This memory requirement 
%the area occupied by the stored ``frozen'' randomness 
a major bottleneck for scaling the system~\cite{SchmuckHardwareOptimizations2019}.
It seems strangely counter intuitive to spend a large amount of memory just to store random vectors.
Thus, our key question is whether collective-state computing can be achieved without memorizing the full array of random vectors. Instead of memorization, can memory requirements be traded off by computation? 

Cellular automata (CA) \cite{Wolfram} are simple discrete computations capable of producing complex random behavior, such as Conway's Game of Life~\cite{GardnerGameOfLife1970}. 
Here we study the randomization behavior of an elementary cellular automaton with rule 90 (CA90) for
%achieving the space-time tradeoff for 
generating distributed representations for collective-state computing. We demonstrate in the context of RC and VSAs that collective-state computing at full performance is possible by storing only short random seed patterns, which are then expanded ``on the fly'' to the full required dimension by running rule   
%We claim that the space-time tradeoff enabled by the elementary 
CA90.  
%comes to rescue as it allows expanding random dense binary distributed representations by replacing the need to memorize ``frozen'' randomness with the need to perform CA90 computations staring from random initial short seeds.  
%The essence of the tradeoff is that on the one hand we will not need to spend large amount of memory for storing the distributed representations but on the other hand we will have to perform CA90 computations. 
%In other words, instead of memorizing the representations in their full length, the use of CA90  allows re-materializing the representations on the fly while using only a fraction of the fully memorized solution at the cost of running CA90 computations every time  when the access to the distributed representations is required. 
%We demonstrate the tradeoff with several common scenarios.
This work is partially inspired by~\cite{YilmazSymbolic2015}, which proposed that RC, VSAs, and CA can benefit from each other, by expanding low-dimensional representations via CA computations into high-dimensional representations that are then used in RC and VSA models.
The specific contributions of this article are:
\noindent
\begin{itemize}
    \item Characterization of the relation between the length of the randomization period of CA90 and the size of its grid;
    \item Analysis of the similarity between CA90 expanded representations in the case when the seed pattern contains errors; 
    \item Experimental evidence that for RC and VSAs the CA90 expanded representations are functionally equivalent to the representations obtained from a standard pseudo-random number generator.
\end{itemize}
\noindent
The article is structured as follows. 
The main concepts used in this study are presented in Section~\ref{sect:concepts}. 
The effect of randomization of states by CA90 is described in Section~\ref{sect:ca:rand}.
The use of RC and VSAs with the CA90 expanded representations is reported in Section~\ref{sect:VSAs:rand}. 
The findings and their relation to the related work are discussed in Section~\ref{sect:discussion}.
%Section~\ref{sect:conclusions} presents the concluding remarks.

%{\color{red}
%This allows us to replace the IM (see Figure 9) by only defining the initial state
%of the CA as a seed pattern and letting it generate the other orthogonal vectors2 for the rest of the channels.
%}

%CA90 expanded representations are applied in 
%The results of experiments confirm that CA90 expanded representations    

%% file: Concepts.tex
\section{Concepts}
\label{sect:concepts}

\subsection{Collective-state computing}
\label{sect:vsas}

As explained in the introduction, collective-state computing 
%is an umbrella term for 
subsumes numerous types of network computations that employ distributed representation schemes based on i.i.d. random vectors. One type is 
%Rather than going through all these methods, we focus on 
VSA or  
%also referred to as 
hyperdimensional computing~\cite{PlateTr, Gallant13, HDNP17}, which has 
been proposed in cognitive neuroscience as a formalism for symbolic reasoning %are a family of bio-inspired models of representing and manipulating concepts 
with distributed representations. Recently, the VSA formalism has been used to formulate other types of 
%as, in our opinion, they are a good formalism for encapsulating various collective-state computing models. 
%models, which are characterized by the use of ``frozen'' randomness and linear superposition to create the collective state. 
%For the sake of brevity, instead of describing many of the models, 
%In the scope of this paper we focus on VSAs as they are a representative collective-state computing model.
%For example, VSAs have recently been linked to other 
collective-state computing models, such as RC~\cite{FradyTheory2018}, compressed sensing~\cite{FradySDR2020}, and randomly connected feed-forward neural networks~\cite{KleykoDensityEncoding2020}. Following this lead, we will formulate the types of collective-state computing,  
%In particular, we introduce several common VSA mechanisms, 
which are used in Section~\ref{sect:VSAs:rand} to study the CA90 expansion. 
%, therefore, the results obtained for VSAs can be easily applied to the other models.  
%In all the models there are variants that use binary random vectors. For example, 
VSAs are defined for different spaces (e.g., real or complex), but here we focus on VSAs with dense binary~\cite{KanervaFully1997} or bipolar~\cite{MAP} vectors where the similarity between vectors is measured by normalized Hamming distance (denoted as $d_h$) for binary vectors or by dot product for bipolar ones (denoted as $d_d$). The VSA formalism will be introduced as we go. 

%VSAs form the collective states by manipulating the atomic vectors.
%To do so, VSAs defines several operations on vectors.
%The key operations are binding, permutation, and bundling.

\subsubsection{Item memory with nearest neighbor search}

A common feature in collective-state computing is that a set of basic concepts/symbols\footnote{Examples of the basic concepts are distinct features in machine learning problems~\cite{Rasanen2015tr, KleykoHolographic2017} or distinct symbols in data structures~\cite{JoshiNgrams2016, HD_FSA, YerxaUCBHD_FSA2018, PashchenkoSubstring2020, KleykoPermuted2016}.} is defined and assigned with i.i.d. random high-dimensional atomic vectors.
In VSAs, these atomic vectors are stored in the so-called item memory (denoted as $\textbf{H}$), which in its simplest form is a matrix with the size dependent on the dimensionality of vectors (denoted as $K$) and the number of symbols (denoted as $D$). The item memory $\textbf{H}$ enables associative or content-based search, that is, for a given query vector $\textbf{q}$ it returns  the nearest neighbor. 
%In the case when the query vector exactly matches one of the vectors $\textbf{H}_i$ stored in the memory, the result of the search is obvious. A more common case, however, is that the query vector is a noisy version of one of the vectors from the item memory.
Given such a noisy query, the memory returns the best match using the nearest neighbor search:  
\noindent
\begin{equation}
%\hat{\textbf{s}}(m-d)=\argmax ( \textbf{W}^{d} \textbf{x}(m)  ),
\underset{i}{\argmin} (d_h(\textbf{H}_i, \textbf{q})).
\label{eq:nn:search}
\end{equation}
\noindent
The search returns the index of the vector that is closest to the query in terms of the similarity metric (e.g., the Hamming distance as in (\ref{eq:nn:search})). 
In VSAs and, implicitly, in most types of collective-state computing, this content-based search is required for selecting and error-correcting results of computations that, in noisy form, have been produced by dynamical transformations of the collective state.  
%read out the result of computation from the noisy collective state. 

\subsubsection{Memory buffer}
\label{sect:mem:bef}

RC is a prominent example of collective-state computing as in echo state networks~\cite{ESN02}, liquid state machines~\cite{LSM02} and state-dependent computation~\cite{buonomano2009state}.
In these models the dynamics of a recurrent network is used to memorize or buffer the structure of a spatio-temporal input pattern. 
In essence, the memory buffer accumulates the input history over time into a compound vector. The recurrent network dynamics attaches time stamps to input arriving at different times, which can be used to later analyze or recover the temporal structure from the compound vector describing the present network state.
%which stores vectors corresponding to symbols from some random sequence of input symbols.
It has been recently shown how the memory buffer task can be analyzed using the VSA formalism~\cite{FradyTheory2018}, which builds on  earlier VSA proposals of the memory buffer task under the name trajectory association task~\cite{PlateBook}.  

Here we use a simplistic variant of the echo state network~\cite{LukoseviciusPracticalESN2012}, called integer echo state network~\cite{KleykointESN2017}.
The memory buffer involves the item memory and two other VSA operations: permutation and bundling. 
The item memory contains a random binary/bipolar vector assigned for each symbol from a dictionary of symbols of size $D$. 
A fixed permutation (rotation) of the elements of the vector (denoted as $\rho$)\footnote{
The cyclic shift is used frequently due to its simplicity.
}
is used to represent the position of a symbol in the input sequence. 
In other words, the permutation operation is used as a systematic transformation of a symbol as a function of its serial position.
For example, a symbol \textit{a} represented by $\textbf{a}$ is associated with its position $i$ in the sequence by the result of permutation (denoted as $\textbf{r}$) as: 
\noindent
\begin{equation}
\label{eq:perm} 
\textbf{r} = \rho^i( \textbf{a}).  
\end{equation}
\noindent
where $\rho^i(*)$ denotes that some fixed permutation $\rho()$ has been applied $i$ times. 
%Similar to the binding operation, the resultant vector $\textbf{r}$ is dissimilar to $\textbf{x}$. 

%In this article, we consider two ways of implementing the limited bundling operation.

The bundling operation forms a linear superposition of several vectors, which in some form is present in all collective-state computing models.
Its simplest realization is an element-wise addition. 
However, when using the element-wise addition, the vector space becomes unlimited, therefore, it is practical to limit the values of the result.
In general, the normalization function applied to the result of superposition is denoted as $f_n(*)$.\footnote{
In the case of dense binary VSAs, the arithmetic sum-vector of two or more vectors is thresholded back to binary space vector by using the majority rule/sum (denoted as $f_m(*)$) where ties are broken at random.
}
The vector $\textbf{x}$ resulting from the bundling of several vectors, e.g.,
\noindent
\begin{equation}
\label{eq:bundle} 
\textbf{x} = f_n(\textbf{a}+\textbf{b} + \textbf{c})
\end{equation}
\noindent
is similar to each of the bundled vectors, which allows storing information as a superposition of multiple vectors~\cite{FradyTheory2018}. 
Therefore, in the context of the memory buffer, the bundling operation is used to update the buffer with new symbols.

The memory buffer task involves two stages: memorization and recall, which are done in discrete time steps.
At the memorization stage, at every time step $t$ we add a vector $ \mathbf{H}_{\textbf{s}(t)}$ representing the symbol $\textbf{s}(t)$  from the sequence $\textbf{s}$ to the current memory buffer $\textbf{x}(t)$, which is formed as: 
\noindent
\begin{equation}
\textbf{x}(t)= f_n ( \rho(\textbf{x}(t-1)) +  \mathbf{H}_{\textbf{s}(t)}  ),
\label{eq:buffer}
\end{equation}
\noindent
where $\textbf{x}(t-1)$ is the previous state of the buffer. 
Note that the symbol added $d$ step ago is represented in the buffer as  $\rho^{d-1}( \mathbf{H}_{\textbf{s}(t-d)})$. 

At the recall stage, at every time step we use $\textbf{x}(t)$ to retrieve the prediction of the delayed symbol stored $d$ steps ago ($\hat{\textbf{s}}(t-d)$) using the readout matrix ($\textbf{W}^{d}$) for particular $d$ using the nearest neighbor search:
\noindent
\begin{equation}
\hat{\textbf{s}}(t-d)=  \underset{i}{\argmax} ( d_d( \textbf{W}_i^{d}, \textbf{x}(t))),
\label{eq:recall:text}
\end{equation}
\noindent
Due to the use of a normalization function $f_n(*)$, the memory buffer possesses the recency effect, therefore, the average accuracy of the recall is higher for smaller values of delay. 
There are several approaches to form the readout matrix and choose a normalization function. 
Please see Appendix~\ref{sect:intesn} for additional details.

\subsubsection{Factorization with resonator network}
\label{sect:fac:rn}

General symbolic manipulations with VSA require one other operation, in addition to bundling, permutation and item memory.
%It is a common requirement to be able 
The represention of an association of two or more symbols, such as a role-filler pair, is achieved by a binding operation, which associates several vectors (e.g., $\textbf{a}$ and $\textbf{b}$) together and produces another vector (denoted as $\textbf{z}$) of the same dimensionality:  
\noindent
\begin{equation}
\label{eq:bind} 
\textbf{z} = \textbf{x}  \oplus \textbf{y},
\end{equation}
\noindent
where the notation $\oplus$ denotes element-wise XOR used for the binding in dense binary VSAs. 
While bundling leads to a vector which is correlated with each of its components, in binding 
%An important property of the binding operation is that, in contrast to the bundling operation, the 
the resulting vector $\textbf{z}$ is pseudo-orthogonal to the component vectors.
Another important property of binding is that it is conditionally invertible. Given all but one components, one
can simply compute from the binding representation of the unknown factor, 
%Nevertheless, when one of the arguments is given, the task of factoring one of the components of $\textbf{z}$ is relatively simple as, 
e.g, $\textbf{z} \oplus \textbf{x} = \textbf{x} \oplus \textbf{x}  \oplus \textbf{y} =  \textbf{y}$.

If none of the factors are given, but are contained in the item memory, the unbinding operation is still feasible but becomes a combinatorial search problem, whose complexity grows exponentially with the number of factors. This problem often occurs in symbolic manipulation problems, for example, in finding the position of a given item in a tree structure~\cite{ResPart1}.
Let us assume that each component (factor; denoted as $\mathbf{f}_i$) comes from a separate item memory ($\prescript{1}{}{\mathbf{H}}, \prescript{2}{}{\mathbf{H}}, ...$), which is called factor item memory, e.g., a general example of a vector with four factors is:
\noindent
\begin{equation}
\label{eq:factors} 
\mathbf{f}_1  \oplus \mathbf{f}_2  \oplus \mathbf{f}_3  \oplus \mathbf{f}_4.
\end{equation}
\noindent
Recent work~\cite{KentResonatorNetworks2019} proposes an elegant mechanism called the resonator network to address the challenge of factoring. 
In the nutshell, the resonator network~\cite{KentResonatorNetworks2019} is a novel recurrent neural network design that uses VSAs principles to solve combinatorial optimization problems. 

To factor the components from the input vector $\mathbf{f}_1  \oplus \mathbf{f}_2  \oplus \mathbf{f}_3  \oplus \mathbf{f}_4$ representing the binding of several vectors, the resonator network uses several populations of units,  $\mathbf{\hat{f}}_1(t), \mathbf{\hat{f}}_2(t), ...$, each of which tries to infer a particular factor from the input vector. 
Each population, called a resonator, communicates with the input vector and all other neighboring populations to invert the input vector using the following dynamics:
\noindent
\begin{equation}
\begin{split}
&\mathbf{\hat{f}}_1(t+1)= f_n \Big( \prescript{1}{}{\mathbf{H}} \prescript{1}{}{\mathbf{H}}^\intercal (\textbf{z} \oplus  \mathbf{\hat{f}}_2(t)  \oplus  \mathbf{\hat{f}}_3(t) \oplus  \mathbf{\hat{f}}_4(t) ) \Big) \\
&\mathbf{\hat{f}}_2(t+1)= f_n \Big( \prescript{2}{}{\mathbf{H}} \prescript{2}{}{\mathbf{H}}^\intercal (\textbf{z} \oplus  \mathbf{\hat{f}}_1(t)  \oplus  \mathbf{\hat{f}}_3(t) \oplus  \mathbf{\hat{f}}_4(t) ) \Big) \\
&\mathbf{\hat{f}}_3(t+1)= f_n \Big( \prescript{3}{}{\mathbf{H}} \prescript{3}{}{\mathbf{H}}^\intercal (\textbf{z} \oplus  \mathbf{\hat{f}}_1(t)  \oplus  \mathbf{\hat{f}}_2(t) \oplus  \mathbf{\hat{f}}_4(t) ) \Big) \\
&\mathbf{\hat{f}}_4(t+1)= f_n \Big( \prescript{4}{}{\mathbf{H}} \prescript{4}{}{\mathbf{H}}^\intercal (\textbf{z} \oplus  \mathbf{\hat{f}}_1(t)  \oplus  \mathbf{\hat{f}}_2(t) \oplus  \mathbf{\hat{f}}_3(t) ) \Big) \\
\end{split}
\label{eqn:resnet:text}
\end{equation}
\noindent
Note that the process is iterative and progresses in discrete time steps, $t$.
In essence, at time $t$ each resonator $\mathbf{\hat{f}}_i(t)$ can hold multiple weighted guesses for a vector from each factor item memory through the VSAs principle of superposition, which is used for the bundling operation. 
Each resonator also uses the current guesses for factors from other resonators. 
These guesses from the other resonators are used to invert the input vector and infer the factor of interest in the given resonator.
The principle of superposition allows a population to hold multiple estimates of factor identity and test them simultaneously. 
The cost of superposition is a crosstalk noise. 
The inference step is, thus, noisy when many guesses are tested at once. 
However, the next step is to use factor item memory $\prescript{i}{}{\mathbf{H}}$  to remove the extraneous guesses that do not fit. 
Thus, the guess $\mathbf{\hat{f}}_i$ for each factor is cleaned-up by constraining the resonator activity only to the allowed atomic vectors stored in $\prescript{i}{}{\mathbf{H}}$.
Finally, a regularization step (denoted as $f_n(*)$) is needed.
Successive iterations of this inference  and clean-up procedure (\ref{eqn:resnet}), eliminate the noise as the factors become identified and find their place in the input vector. 
When the factors are fully identified, the resonator network reaches a stable equilibrium and the factors can be read out from the stable activity pattern.
Please refer to Appendix~\ref{sect:resnet} for additional motivation and explanation of the resonator network.

\subsection{Cellular automata-based expansion}
\label{sect:ca}

The CA is a discrete  computational model consisting of a regular grid of cells~\cite{Wolfram} of size $N$. 
Each cell can be in one of a finite number of states (the elementary CA is binary). 
States of cells evolve in discrete time steps according to some fixed rule. 
In the elementary CA, the new state of a cell at the next step depends on its current state and the states of its immediate neighbors.
Despite the seeming simplicity of the system, amongst the elementary CAs there are rules (e.g., rule 110) that make CA dynamics operate at the edge of chaos, and which were proven to be Turing complete~\cite{Cook2004}. 
In the scope of this article, we consider another rule -- rule 90 (CA90) as it possess several properties highly relevant for collective-state computing.
%Also, notation $CA^n_{90}$ will be used to denote the $n$-th computing step of the CA. 

\begin{figure}[tb]%[!ht]%[t!]
\centering
\includegraphics[width=1.0\columnwidth]{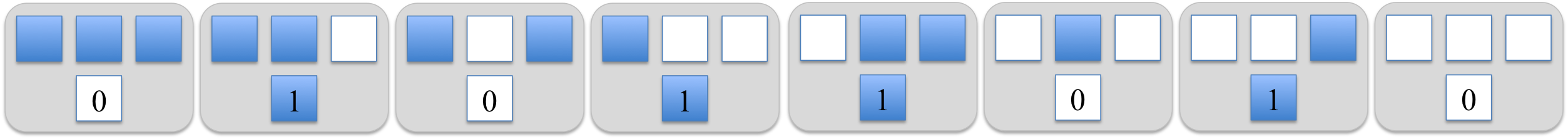}
\caption{The assignment of new states for a center cell when the CA uses rule 90. A hollow cell corresponds to zero state while a shaded cell marks one state.
}
\label{fig:rule90}
\end{figure}

\begin{figure}[tb]%[!ht]%[t!]
\centering
\includegraphics[width=1.0\columnwidth]{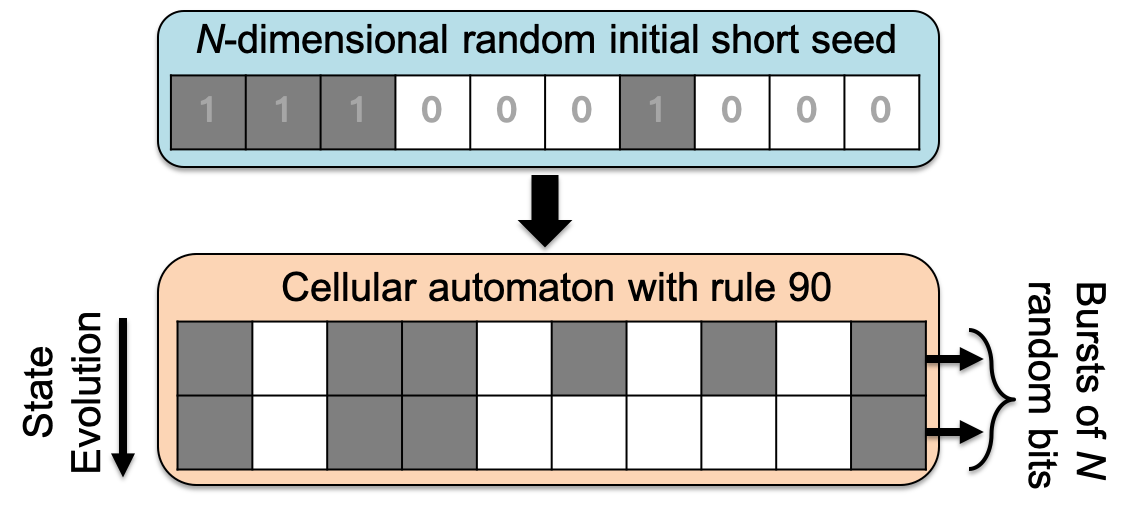}
\caption{
The basic scheme for expanding distributed representations with CA90 from some initial short seed.
The dimensionality of the seed $N$ equals to the the size of the CA grid.
}
\label{fig:CA90:expand}
\end{figure}

%The CA performs computations locally. 
%The new state of a cell is determined by previous states of the cell itself and its two neighboring cells (left and right). 
%In other words, 
In the elementary CA there are only three cells involved in a  computation step, and for CA with binary states, there are in total $2^3=8$ possible input combinations for each input there are two possible assignments for the output cell, which makes in total $2^8 =256$ combinations where each particular assignment defines a rule. 
%A rule assigns new states for each of these combinations.  
Fig.~\ref{fig:rule90} shows all input combinations and corresponding assignment of output states for
CA90. 
CA90 assigns  the next state of a central cell based on the previous states of the neighbors. 
In particular, the new state is the result of XOR operation on the states of the
neighboring cells.  
This is particularly attractive because CA90 has a computational advantage since the CA
implementation can be easily vectorized and implemented in hardware (especially when working with cyclic boundary conditions\footnote{Cyclic boundary condition means that the first and the last cells in the grid are considered to be neighbors.
}). 
For example, if at time step $t$ vector $\textbf{x}(t)$ describes the states of the CA grid,  then the grid state at $t+1$ is computed as: 
\noindent
\begin{equation}
\label{eq:ca90}
\textbf{x}(t+1)= \rho^{+1}(\textbf{x}(t)) \oplus \rho^{-1}(\textbf{x}(t)),
\end{equation}
\noindent
where $\rho^{\{+1, -1\}}$ is the notation for cyclic shift to the right or left by one.

Since in the context of this study we use CA90 for the purposes of randomization, we will call the state of the grid $\textbf{x}(0)$ at the beginning of computations as an initial short seed. 
% is this really worth pointing out?
It is worth pointing out that CA90 formulated as in (\ref{eq:ca90}) is a sequence of VSA operations~\cite{Gayler03}. 
Given the vector $\textbf{x}$ as an argument, by performing two rotations ($\rho^{+1}(\textbf{x})$ and $\rho^{-1}(\textbf{x})$) and then binding the results of rotations together ($\rho^{+1}(\textbf{x}) \oplus \rho^{-1}(\textbf{x})$), we implement CA90.

%$\textbf{x}(t+1)= \text{Sh}(\textbf{x}(t),1) \oplus \text{Sh}(\textbf{x}(t),-1)$, where $\text{Sh}()$ is the notation for cyclic shift operation

The core idea of this paper is to use CA90 to generate a distributed representation of expanded dimensionality that can be used within the context of collective-state computing. This expansion must have certain randomization properties and be robust to perturbations.
Fig.~\ref{fig:CA90:expand} presents the basic idea of obtaining an expanded dense binary distributed representation from a short initial seed. In essence, the seed is used to initialize the CA grid.
Once initialized, CA90 computations are applied for $L$ steps.
At every step of the evolution, the state of the grid provides a new burst of $N$ bits, which can be either used on the fly (without memorization) to make the necessary manipulations and then erased, or concatenated (with memorization) to the previous states if the distributed representation should be re-materialized explicitly. 
In any case, the dimensionality of the expanded representation is $K=NL$.

\subsection{CA90 and VSAs}

Section~\ref{sect:discussion} will present the joint use of RC, VSAs, and CA90 expansion. 
Amongst the related works discussed,~\cite{KleykoBrainsCA2017} is the most relevant, as it uses the randomization property of CA90.
In particular, this work identified the following useful properties of CA90 for VSAs: 
\begin{enumerate}
\item Random projection; 
\item Preservation of the binding operation; 
\item Preservation of the cyclic shift.
\end{enumerate}

By random projection we mean that when CA90 is initialized with a random
state $\textbf{x}(0)$ ($p_1 \approx p_0 \approx 0.5$), which should be seen as an initial short seed, its evolved state at step $t$ is a vector $\textbf{x}(t)$ of the same size and density. 
Moreover, during the randomization period (see Section~\ref{sect:ca:rand:el}) $\textbf{x}(t)$ is dissimilar to the initial short seed $\textbf{x}(0)$, i.e., $d_h(\textbf{x}(0),\textbf{x}(t)) \approx 0.5$ as well as to the other states in the evolution of the seed.

\begin{figure}[tb]%[!ht]%[t!]
\centering
\includegraphics[width=1.0\columnwidth]{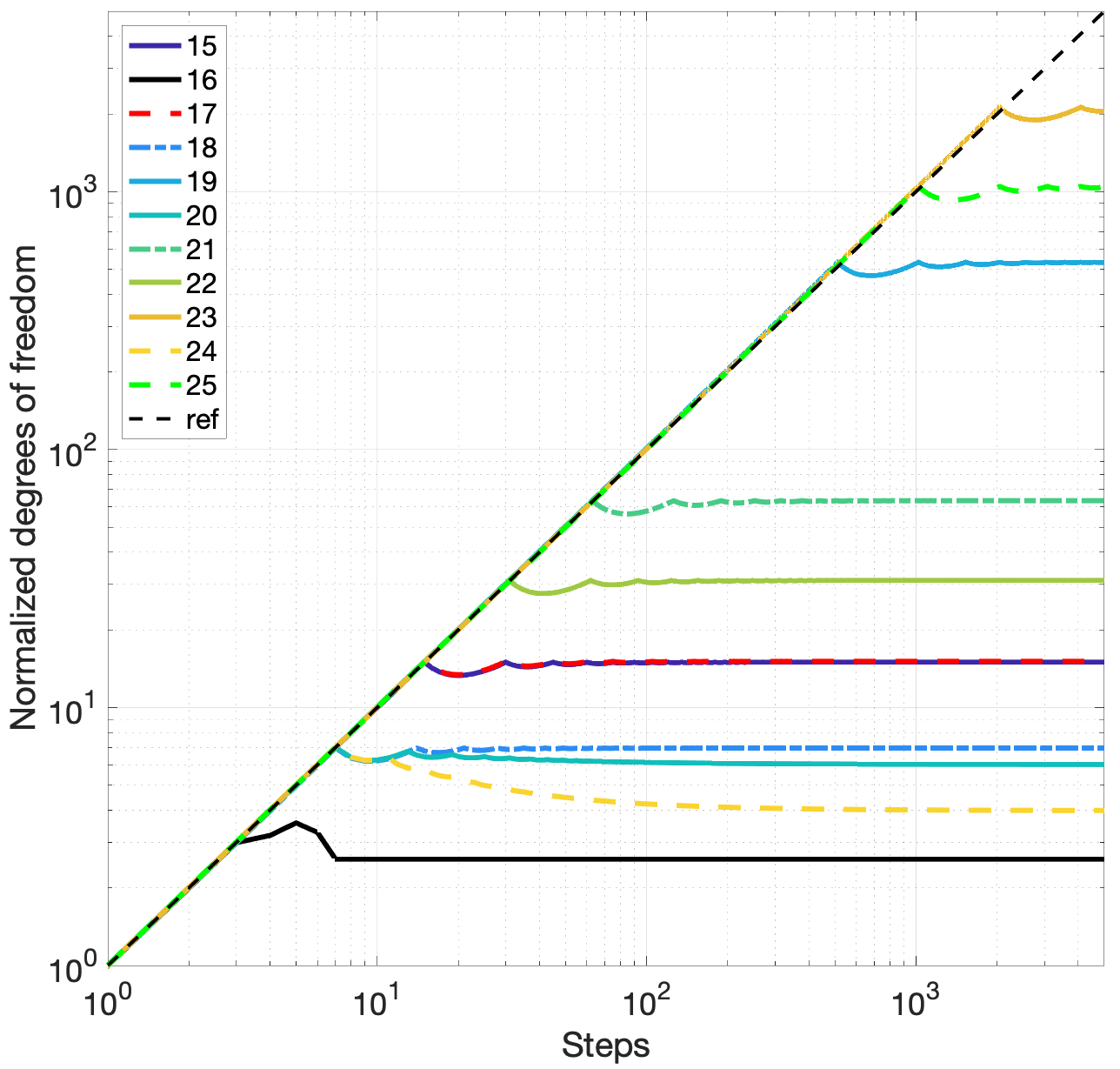}
\caption{
The normalized degrees of freedom for different values of the grid size of CA90. 
The evolution of CA90 is reported for $5,000$ steps.
The number of short seeds in the item memory was fixed to $100$. 
The reported values were averaged over $100$ simulations randomizing initial short seeds.
Note the logarithmic scales of the axes.
}
\label{fig:ca:period:example}
\end{figure}

The preservation of the binding operation refers to the fact that if a seed $\textbf{c}(0)$ is the result of the binding of two other seeds: $\textbf{c}(0)= \textbf{a}(0) \oplus \textbf{b}(0)$ then after $t$ 
computational steps of  CA90, the evolved state $\textbf{c}(t)$ can be obtained by binding the evolved states of the initial seeds $\textbf{a}(0)$ and $\textbf{b}(0)$ used to form  $\textbf{c}(0)$, i.e., $\textbf{c}(t)= \textbf{a}(t) \oplus \textbf{b}(t)$.

\begin{figure}[tb]%[!ht]%[t!]
\centering
\includegraphics[width=1.0\columnwidth]{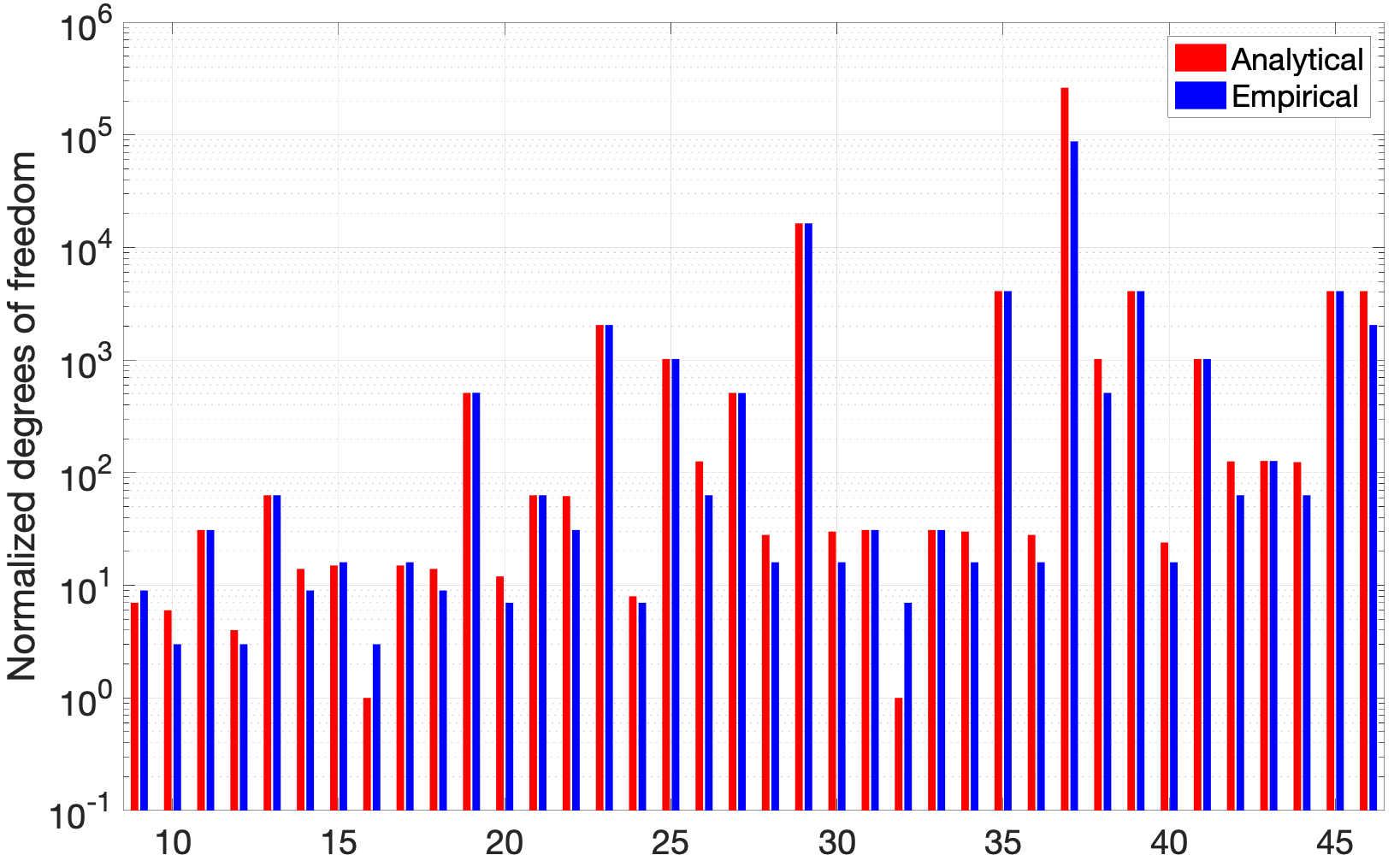}
\caption{
The empirically measured randomization period (blue) and the analytical periodic cycles~\cite{MartinPropertiesCA1984} (red) for the grid size in the range $[9, 46]$.
Note the logarithmic scale of  \textit{y}-axis.
}
\label{fig:ca:periods}
\end{figure}

Finally, CA90 computations  preserve a special case of the  permutation operation -- cyclic shift by $i$ cells. 
Suppose $\textbf{d}(0)=\rho^i(\textbf{a}(0))$ is an initial seed.
After $t$ computational steps of CA90, the cyclic shift of the  evolved seed $\textbf{a}(t)$ by $i$ cells equals the evolved  shifted seed $\textbf{d}(t)$ so that $d_h(\textbf{d}(t), \rho^i(\textbf{a}(t))=0$.

%These properties of CA90 constitute the core of the memory cloning pipeline presented in the next section.
% Rule 90 has several  properties highly relevant for the task of mapping a system
% of vectors into  unrelated part  of the hyperdimensional space while
% preserving internal relationships between  vectors. First, when rule 90
% starts from a random initial states (i.e., with an vector),  its evolved
% states form random vector ($p_1 \approx p_0 \approx 0.5$),  which is
% unrelated to the initial vector, at each computational step.  In other words,
% given an vector as an initial configuration, the cellular  automaton with
% rule 90 maps it to a different part of the hyperdimensional space.

%% file: CA90_Randomization.tex
%\newpage

\section{Randomization of states by cellular automaton}
\label{sect:ca:rand}

%FS: I did not understand this sentence: "Next, in terms of length of randomization period are even grid sizes followed by odd sizes." Explanation could also be improved in the result section

\subsection{Errorless randomization}
\label{sect:ca:rand:el}

Usually, distributed representations in collective-state computing use i.i.d. random vectors.
Similarly, we start with i.i.d. random vectors for short seeds.
However, in contrast to the conventional approach, we are going to expand the dimensionality of the seed vectors via CA90 with boundary conditions. 
An important question for expansion is what are the limits of CA90 in terms of producing randomness?

One very useful empirical tool for answering this question is calculation of degrees of freedom from the statistics of normalized Hamming distances between binary vectors (see, e.g.,~\cite{Daugman2003} for an example).
Given that $p_h$ denotes the average normalized Hamming distance and $\sigma_h$ denotes its standard deviation, the degrees of freedom are calculated as: 
\noindent
\begin{equation}
F= p_h(1-p_h)/\sigma_h^2.
\label{eq:deg:fre}
\end{equation}
\noindent
Due to the randomization properties of CA90, we expect that after a certain number of steps it will produce new degrees of freedom. 
To be able to compare different grid sizes, we will report the degrees of freedom normalized by the grid size, i.e., $F/N$. In other words, if a single step of CA90 increased $F$ by $N$ (best case), the normalized value would increase by $1$.

Fig.~\ref{fig:ca:period:example} presents the normalized degrees of freedom measured for $5,000$ steps of CA90 for different grid sizes using the same values as in~\cite{Wolfram} (p. 259).
From the figure we can draw several interesting observations. 
First for all of the considered grid sizes, the degrees of freedom grows linearly at the beginning (following the reference, dashed black line, which indicates degrees of freedom in random binary vectors of the corresponding size). 
At some point, however, the degrees of freedom reaches a maximum value and starts to saturate, as we would expect. 
We are interested in the period of linear growth, and we call this the randomization period.
Second, we see that larger grid sizes typically have longer randomization periods.
For example, the longest randomization period of $2,047$ steps was observed for $N=23$ (but this is not the largest grid size).
Third, the randomization period of odd grid sizes are always longer than that of the even ones. 
For example, the randomization periods for $N=22$ and $N=24$ were only $31$ and $7$, respectively (cf. $2,047$ for $N=23$). Thus, there is a dependency between $N$ and the randomization period, but, despite the above observations, there is not a clear general pattern connecting the grid size to the length of the randomization period.

%\todo[inline, color=yellow]{
%Need explanation why behavior of generation is in such case?
%}

The good news, however, is that the length of the randomization period is closely related to the length of periodic cycles (denoted as $\Pi_N$) in CA90 discovered in~\cite{MartinPropertiesCA1984}. 
In short, the irregular behaviour of randomization periods and periodic cycles is a consequence of their dependence on number theoretical properties of $N$;~\cite{MartinPropertiesCA1984} provides the following characterization of periodic cycles $\Pi_N$ in CA90:
\begin{itemize}
    \item For CA90 with $N$ of the form $2^j$, $\Pi_N=1$; %because all activity in CA90 decays to zero after the first $2^{j-1}$ steps.
    \item For CA90 with $N$ even but not of
the form $2^j$, $\Pi_N=2\Pi_{N/2}$;
    \item For CA90 with $N$ odd, $\Pi_N| \Pi_N^*=2^{sord_N(2)}-1$, where $sord_N(2)$ is the multiplicative ``sub-order'' function of 2 modulo $N$, defined as the least integer $j$ such that $2^j = \pm 1 \mod N$. 
\end{itemize}
%When $N$ is of the form $2^j$, $\Pi_N=1$ 

\begin{figure}[tb]%[!ht]%[t!]
\centering
\includegraphics[width=1.0\columnwidth]{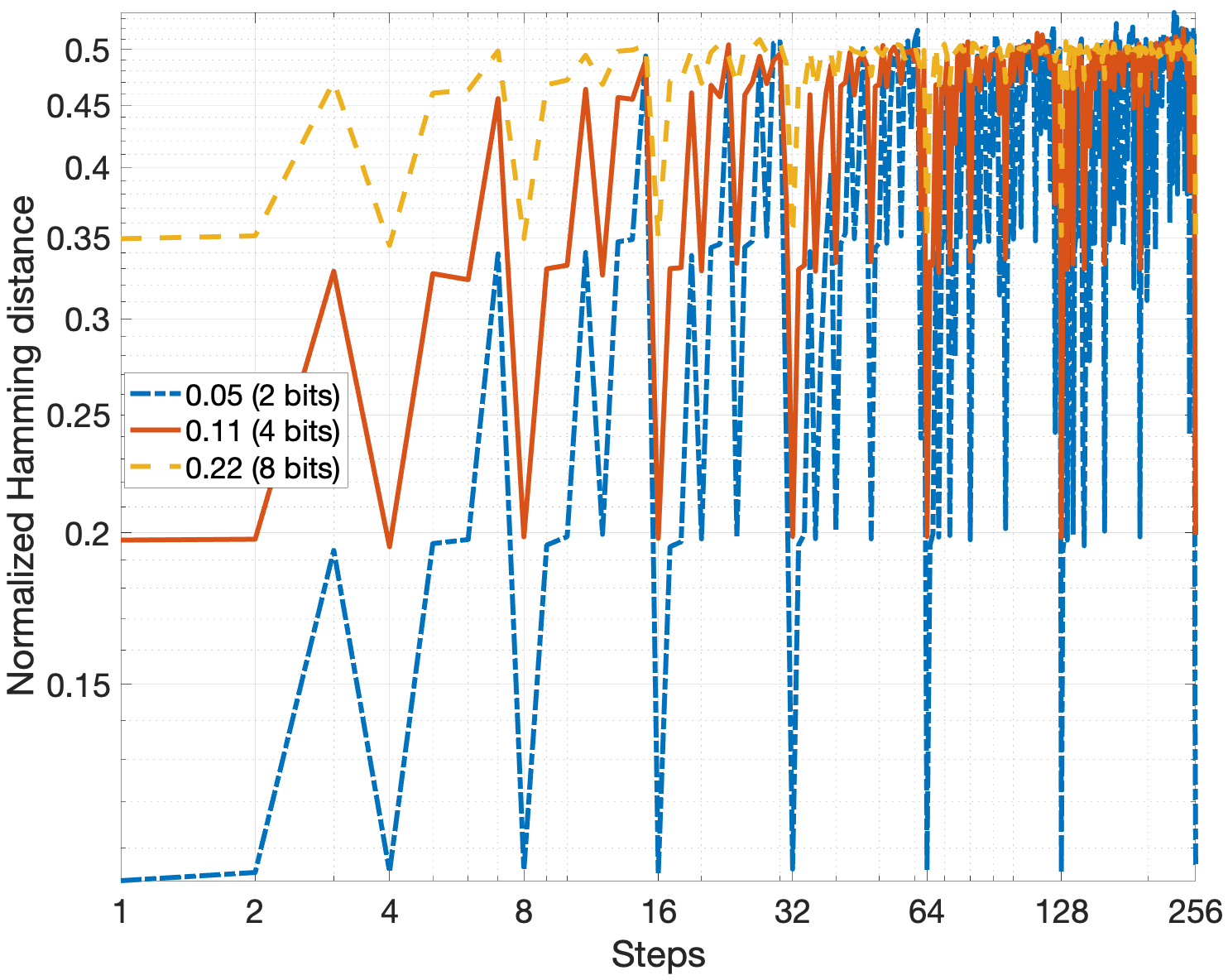}
\caption{
The normalized Hamming distance between the original and noisy vectors obtained from short seeds for $N=37$ during the first $256$ steps of CA90 evolution.
The reported values were averaged over $500$ simulations where both initial short seeds and errors were chosen at random.
Note the logarithmic scales of axes.
}
\label{fig:ber:ca:cur}
\end{figure}

Fig.~\ref{fig:ca:periods} presents the empirically measured randomization periods as well as analytically calculated periodic cycles $\Pi_N^*$~\cite{MartinPropertiesCA1984} for the grid size in the range $[9, 46]$.
First, we see that when $N$ is odd, the randomization period equals to the periodic cycle. 
The only exception is the case when $N=37$, but this is just the first exception where $\Pi_N = \Pi_N^*/3$.
Second, when $N$ is of the form $2^j$, the randomization period does not equal one because the CA90 is producing activity for $2^{j-1}$ steps, which increases the degrees of freedom.  
In fact, the randomization period in this case is $2^{j-2}-1$.
Third, when $N$ is even, the CA90 produces $\Pi_N$ unique grid states but the patterns of Hamming distances  between the states evolved from two random initial short seeds start to repeat after $\Pi_N/2$ steps, thus, they do not contribute new degrees of freedom. 
Therefore, the randomization period is two times lower than the periodic cycle. 
Aggregating these points, with respect to the randomization period of CA90, we have the following: 
\begin{itemize}
    \item For CA90 with $N$ of the form $2^j$, the randomization period is $2^{j-2}-1$;
    \item For CA90 with $N$ even but not of the form $2^j$, the randomization period is $\Pi_{N/2}$;
    \item For CA90 with $N$ odd, the randomization period is divide of $\Pi_N^*=2^{sord_N(2)}-1$. 
\end{itemize}

\subsection{The effect of noise in the short seed}

\begin{figure}[tb]%[!ht]%[t!]
\centering
\includegraphics[width=1.0\columnwidth]{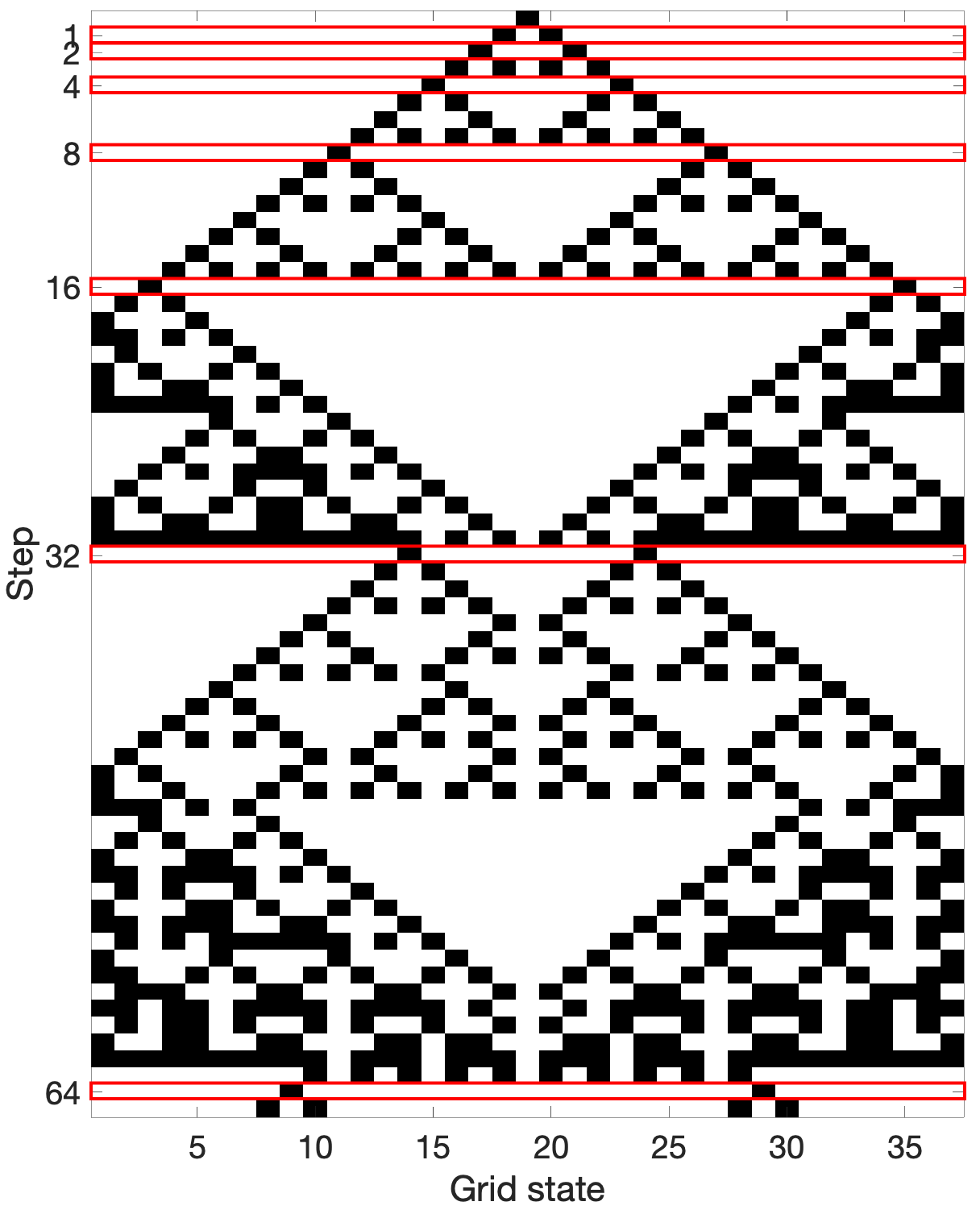}
\caption{
The evolution of CA90 for $65$ steps, $N=37$; the initial state includes one active cell, which can be thought as introducing one bit flip to some random initial short seed.
All steps of the form $2^j$ are highlighted by red rectangles. 
}
\label{fig:ca:1error}
\end{figure}

%\todo[inline, color=yellow]{
%Report behavior of CA90 w.r.t. to errors
%}

In the previous section, we have seen how CA90 can be used to expand initial short seeds into longer randomized representations.
This property could be utilized by a collective-state system for efficient communication by exchanging only short seeds and expanding the seed with CA90.
% A potential use-case for this property of CA90 is when a collective-state system is distributed and a communication is involved. 
% In this case, in order to save communication resources different parts of the system could exchange only short seeds and then generate the longer distributed representations with CA90. 
In such a scenario, one might expect errors when communicating the short seeds, therefore, it is important to understand how the evolution of an expanded distributed representation is affected by errors in the initial short seed.

We address this issue by observing the empirical behavior for $N=37$ and the first $256$ steps of CA90 evolution. 
Fig.~\ref{fig:ber:ca:cur} reports the averaged normalized Hamming distance for an errorless short seed and a noisy version of it, where either $2$ (dash-dot line), $4$ (solid line), or $8$ (dashed line) bits were flipped randomly. 
The results reported are for the corresponding states of the grid at a given step; not for the concatenated states.  
This shows that even a single step of CA90 increases the normalized Hamming distance between the evolved states. 
For example, when $4$ bits were flipped the normalized Hamming distance between the seeds was $4/37\approx0.11$ while after a single step of CA90 it increased to almost $0.2$, almost doubling.
Further, we observe that the normalized Hamming distance will never be lower than after the first step.  

What is very interesting is that the distances induced by errors change in a predictable manner. 
We see that the errors reset to the lowest possible value at regular intervals: each CA90 step of the form $2^j$.
This behavior of CA90 suggests that we can mitigate the impact of errors when expanding short seeds. In order to minimize the distance between the errorless evolutions and their noisy versions, one should only use the CA90 expansion in steps of the form $2^j$, which places additional limits for the possibility of dimensionality expansion.

\begin{figure}[tb]%[!ht]%[t!]
\centering
\includegraphics[width=1.0\columnwidth]{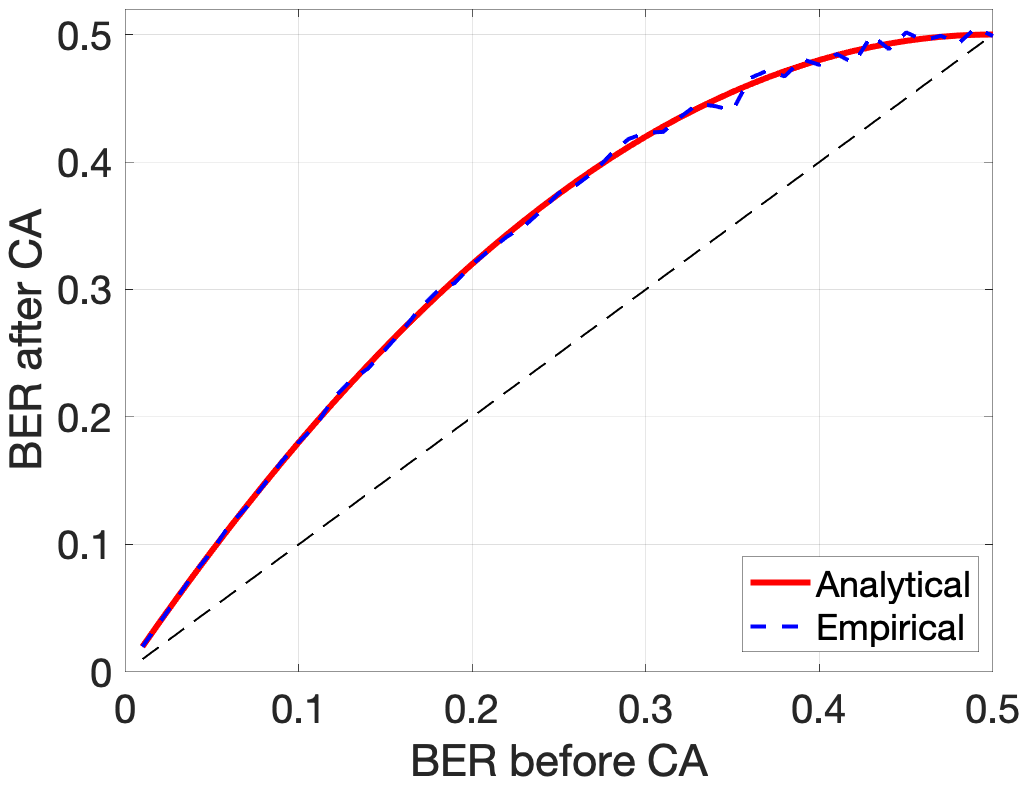}
\caption{
The expected BER for CA90 steps of the form $2^j$ against the BER in the short seeds. 
The solid line is analytical calculation while the dashed line was measured empirically.
The empirical results were averaged over $10$ simulations.
}
\label{fig:ber:ca:tot}
\end{figure}

To understand the observed cyclic behavior of CA90, we examine the case when the initial state of the grid includes only one active cell.
Due to the fact that CA90 is additive, the active cell can be interpreted as one bit flip of error introduced to some random initial short seed.
Fig.~\ref{fig:ca:1error} demonstrates the evolution of the considered configuration for the first $65$ steps.
Red rectangles in Fig.~\ref{fig:ca:1error} highlight the steps of the form $2^j$. At these steps there are only $2$ active cells.
The behavior of the configuration with the single active cell explains both why for small number of bit flips in Fig.~\ref{fig:ber:ca:cur} the normalized Hamming distance approximately doubled after the first step and why the distance reset for every step of the form $2^j$. 
% but why? does this always hold?

% Seems too obvious to mention, unless there is a deeper point?
%Note that when the number of bit flips in a short seed increases (e.g., $8$ in Fig.~\ref{fig:ca:1error}) the number of errors after the first CA90 step is less than doubled.  
%That is because the errors start to interact and might cancel each other after applying CA90.  
%In fact, 
Given the Bit Error Rate (BER, number of bit flips) in the short seed ($p_{bf}$), we can calculate the BER after CA90 expansion (denoted as $p_{CA}$) for steps of the form $2^j$ as follows:
\noindent
\begin{equation}
p_{CA}= 2 p_{bf}^2 (1-p_{bf}) + 2  p_{bf} (1-p_{bf})^2 = 2p_{bf}(1-p_{bf}).
\label{eq:ber:ca}
\end{equation}
\noindent
The intuition here is that due to the local interactions of CA, it is enough to only consider cases as in Fig.~\ref{fig:rule90}.
In particular, we are only interested in cases, which result in active cells at the next step. 
There are only 4 such cases: two with two active cells and two with one active cell; enumerated in (\ref{eq:ber:ca})  

Fig.~\ref{fig:ber:ca:tot} plots the analytical $p_{CA}$ according to (\ref{eq:ber:ca}) against the empirical one obtained in numerical simulations, we see that the curves match.

%% file: Results.tex
%\newpage
%\phantom{text} 
%\newpage

\section{Experimental demonstration of CA90 expansion for collective-state computing}
\label{sect:VSAs:rand}

% This figure is redundant with figure 2.
% \begin{figure}[tb]%[!ht]%[t!]
% \centering
% \includegraphics[width=0.6\columnwidth]{img/CAItemExp}
% \caption{
% The functionality of a circuit making CA90 expansion on the whole item memory. 
% Every step of CA90 expands the dimensionality of vectors by $N$ bits. 
% }
% \label{fig:CA90:expand:item}
% \end{figure}

\begin{figure}[tb]%[!ht]%[t!]
\centering
\includegraphics[width=1.0\columnwidth]{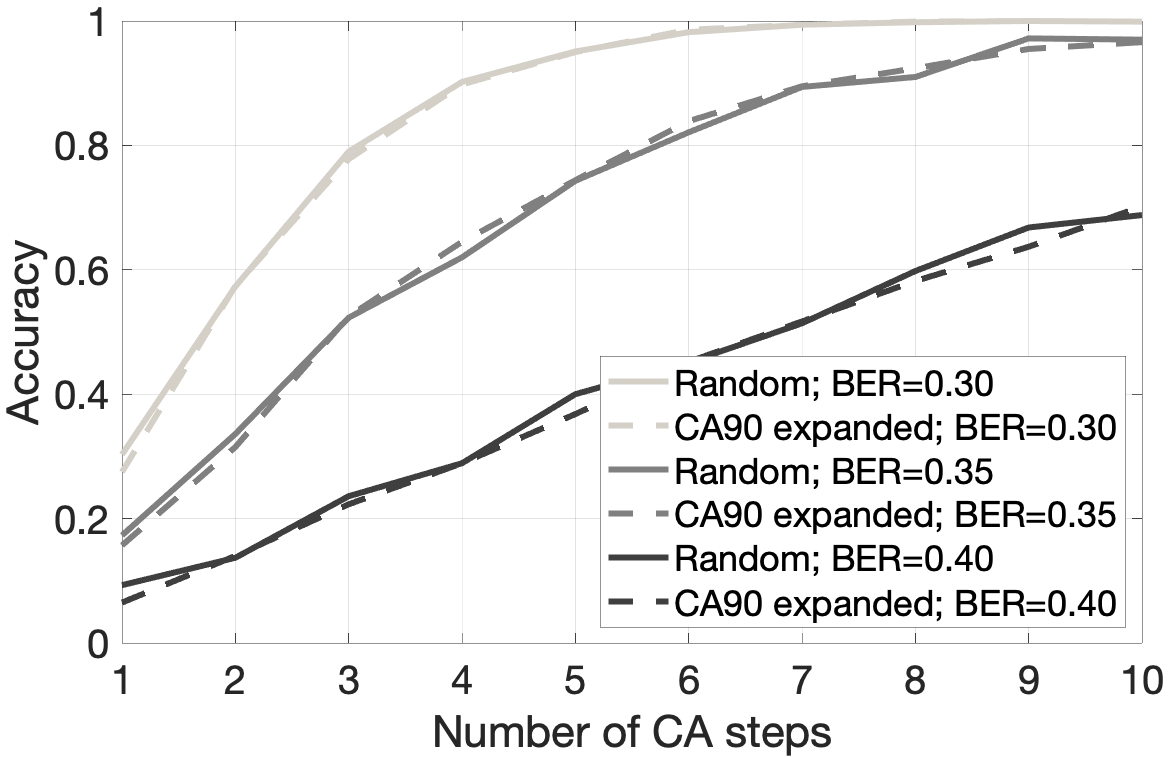}
\caption{
The usage of i.i.d. random vectors against the CA90 expanded representations in the item memory.
The figure reports the accuracy of the nearest neighbor search where a query was a noisy version of one of the vectors stored in the item memory.
The noise was introduced in the form of bit flips.
Three different values of Bit Error Rates were simulated ($\{0.30, 0.35, 0.40\}$).
The dimensionality of the initial short seeds was set to $N=23$.
The size of the item memory was set to $100$.
The reported values were averaged over $1,000$ simulations randomizing initial short seeds, random item memories, and noise added to queries.
}
\label{fig:item:mem:CA}
\end{figure}

This section focuses on using CA90 expansion for RC and VSAs. 
In several scenarios, we provide empirical evidence that expanded vectors obtained via CA90 computations are functionally equivalent to i.i.d. random vectors.\footnote{
In the scope of this article by random vectors we mean vectors generated with the use of a standard pseudo-random number generator. 
So strictly speaking, they should be called pseudo-random vectors but the term random is used to oppose them to the vectors obtained with CA90 computations.
}
%Fig.~\ref{fig:CA90:expand:item} shows how the item memory storing initial short seeds is expanded with CA90 in the experiments reported below. 
%The code for reproducing the results of the experiments  is available as the supplementary materials to this article. 

\subsection{Nearest neighbor search in item memories}

%\todo[inline, color=yellow]{
%Single neuron vs. CA neuron.
%Input snippet and output snippet.
%Add expansion provide between inner product
%kNN neuron dot product with its stored synapse. 
%}

One potential application of CA90 expansion of vectors will be for ``on the fly'' generation of item memories as used in RC and VSAs.
An item memory is used to decode the output of a collective-state computation, where often the nearest neighbor to a query vector within the item memory is to be found.
As mentioned before, when there is noise in the query vector, the outcome of the search may not always be correct. 
Therefore, we explored the accuracy of the nearest neighbor search when the query vector was significantly distorted by noise. We compare two item memories: one with i.i.d. random vectors, and the other with CA90 expanded vectors where only initial short seeds ($N=23$) were i.i.d. random.
Fig.~\ref{fig:item:mem:CA} reports the accuracy results of simulation experiments. 
The item memory with vectors based on CA90 expansion demonstrated the same accuracy as the item memory with fully i.i.d. random vectors. 
%This result provides us the initial empirical evidence that CA90 expanded representations can be used in item memories without sacrificing the performance of the system. 

%\newpage
\subsection{Memory buffer}
\label{sec:mem:buf}

\begin{figure}[tb]%[!ht]%[t!]
\centering
\includegraphics[width=1.0\columnwidth]{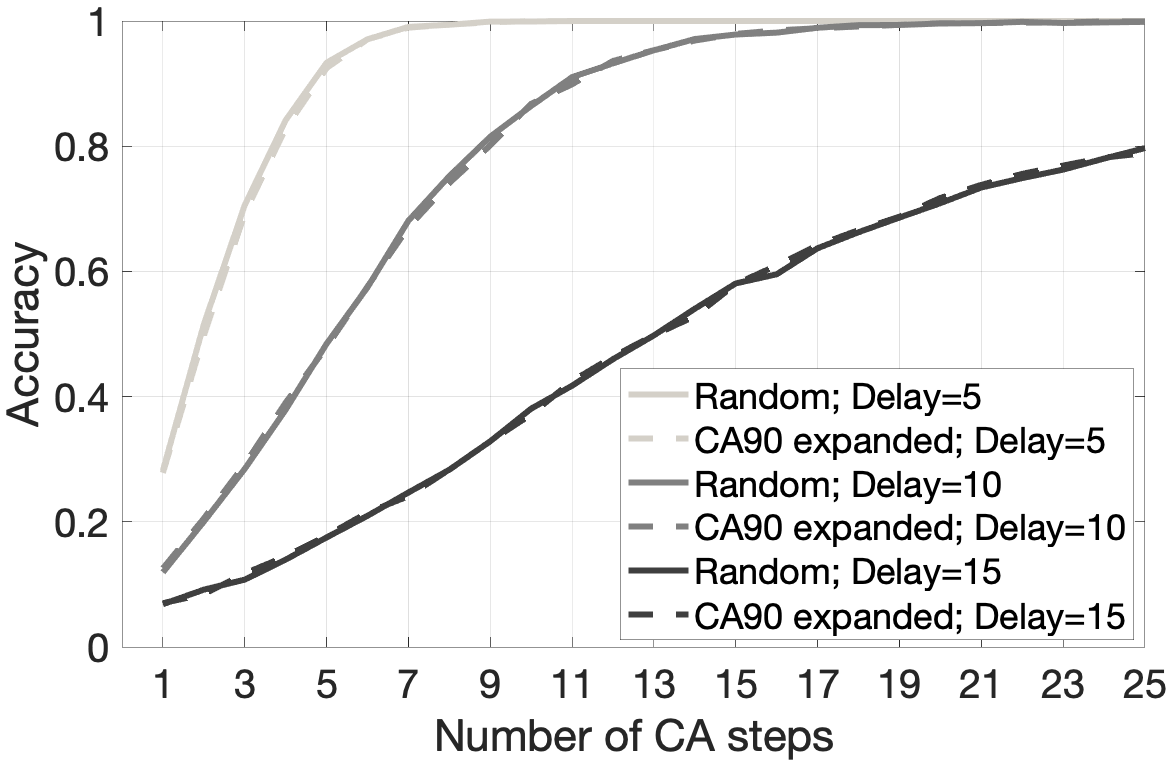}
\caption{
The usage of i.i.d. random vectors against the CA90 expanded representations in the memory buffer task; $D=27$ in the experiments.
The figure reports the accuracy of the  correct recall of symbols for three different values of delay ($\{5, 10, 15\}$).
The dimensionality of the initial short seeds was set to $N=37$.
The reported values were averaged over $10$ simulations randomizing initial short seeds, random item memories, and traces of symbols to be memorized.
}
\label{fig:mem:buf:CA}
\end{figure}

\begin{figure*}[tb]%[!ht]%[t!]
\centering
\includegraphics[width=1.95\columnwidth]{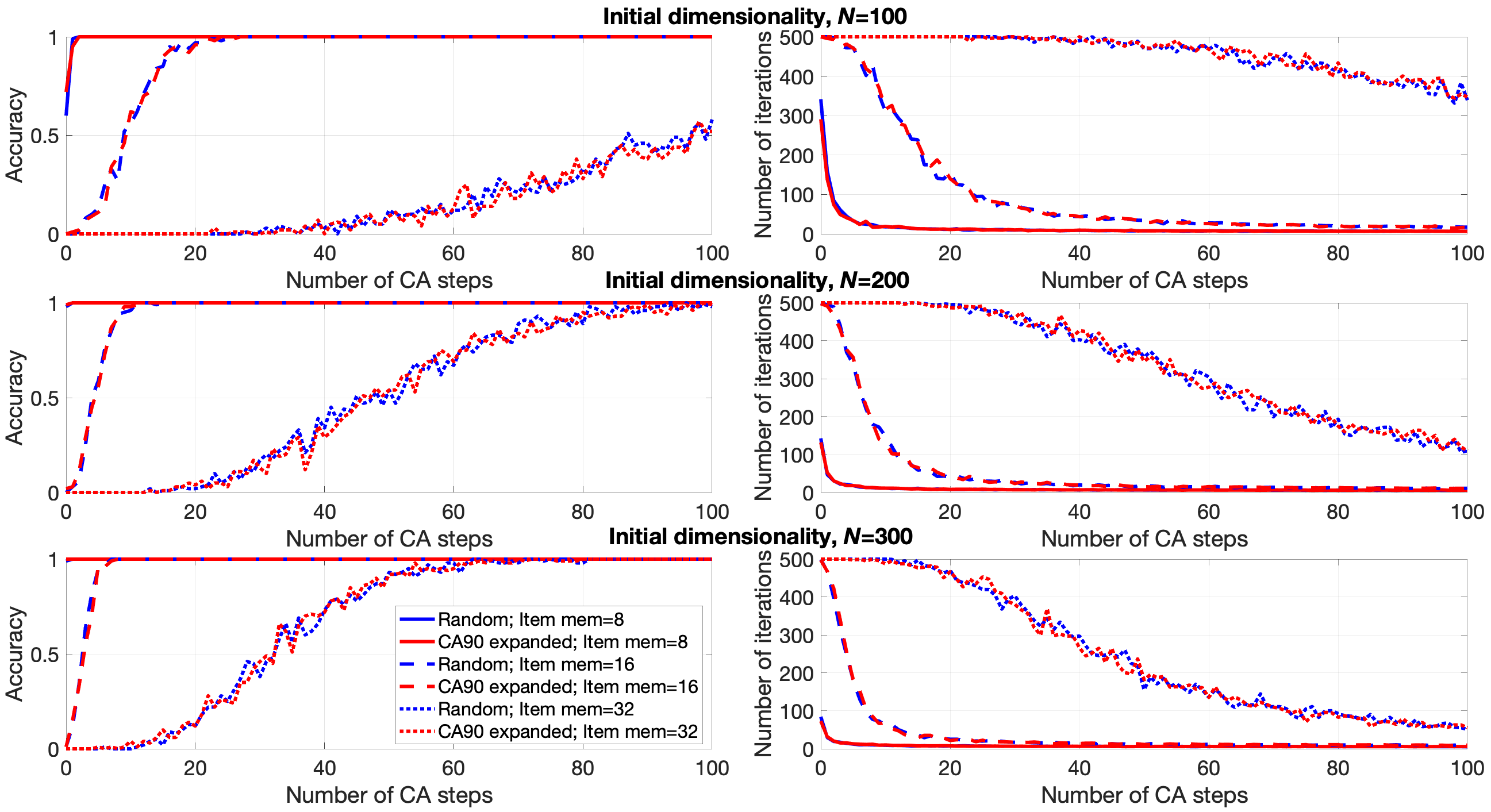}
\caption{
The usage of random vectors against CA90 expanded representations in the resonator network. 
Left column reports the average accuracies while right column reports the average number of iterations until convergence. 
The maximal number of iterations was set to $500$.
The dimensionality of the initial short seeds varied between $\{100, 200, 300\}$.
The evolution of CA90 is reported for the first $100$ steps.
The size of an individual item memory varied between $\{8, 16, 32\}$.
The number of factors was fixed to four. 
The reported values were averaged over $100$ simulations randomizing initial short seeds.
}
\label{fig:res:error:less}
\end{figure*}

Next, we demonstrate the use of CA90 expanded representations in the memory buffer task described in Section~\ref{sect:mem:bef}. 
In these experiments, we measured the accuracy of recall from the memory buffer when the item memory was created from initial short seeds ($N=37$) by concatenating the results of CA90 computations for several steps, $K=NL$. 
%Thus, the dimensionality of the vectors in the expanded item memory was equal to the dimensionality of the initial short seeds multiplied by the number of CA steps. 
As a benchmark, we used an item memory with i.i.d. random vectors of matching dimensionality.
Three different values of delay were considered: $\{5, 10, 15\}$.
The results are reported in Fig.~\ref{fig:mem:buf:CA}.
As expected, we observe that increasing the dimensionality of the memory buffer increased the accuracy of the recall. 
The main point, however, is that the memory buffer made from CA90 expanded representations demonstrates the same accuracy as the memory buffer made from i.i.d. random vectors.

\subsection{Resonator network factoring in the error-free case}
\label{sec:res:resnet:el}
%\todo[inline, color=yellow]{
%Report results for resonator versus random
%}

To further assess the quality of vectors obtained via the results of CA90 computations, we also examined their use in the resonator network~\cite{ResPart1}. 
%performed the comparison with i.i.d. random vectors of the matching dimensionality on the resonator network. 
Please see Section~\ref{sect:fac:rn} and  Appendix~\ref{sect:resnet} for details of the resonator network. 

It is important to emphasize that due to the preservation of the binding operation by CA90, that multiple aspects of the resonator network can benefit from CA90 expansion. Both the composite input vector and the factor item memories do not have to be memorized explicitly, but rather can be expanded from seeds.
The factorization process is computed at each level of CA90 expansion, with the vector dimensions increasing by $N$ for each CA step.
The outputs of the resonator network are collected and compared to the ground truth, and averaged over many randomized simulation experiments.
Fig.~\ref{fig:res:error:less} presents the average accuracies (left column) and the average iterations until convergence (right column) for three different dimensionalities of initial short seeds $\{100, 200, 300\}$ and three sizes of factor item memories $\{8, 16, 32\}$. 
The number of factors was set to four. 
The simulations considered the first $100$ steps of CA90, which was less than the randomization period ($1,023$) of the shortest seed ($N=100$). 

The performance of the resonator network was as expected.
For a given dimensionality of short seed and item memory size, the average accuracy increased with the increased number of CA90 steps -- as practically it means using vectors of larger dimensionality. 
The number of iterations in contrast decreased for larger vectors. 
%As expected, the larger size of item memories lead to higher accuracies and lower number of iterations.
Importantly, there was no notable difference in the performance of the resonator network both in terms of the accuracy and number of iterations when operating with CA90 expanded representations. 
This further confirms that it is reasonable to use CA90 expanded representations in order to trade-off memory for computation.

%\newpage

\begin{figure*}[tb]%[!ht]%[t!]
\centering
\includegraphics[width=2.0\columnwidth]{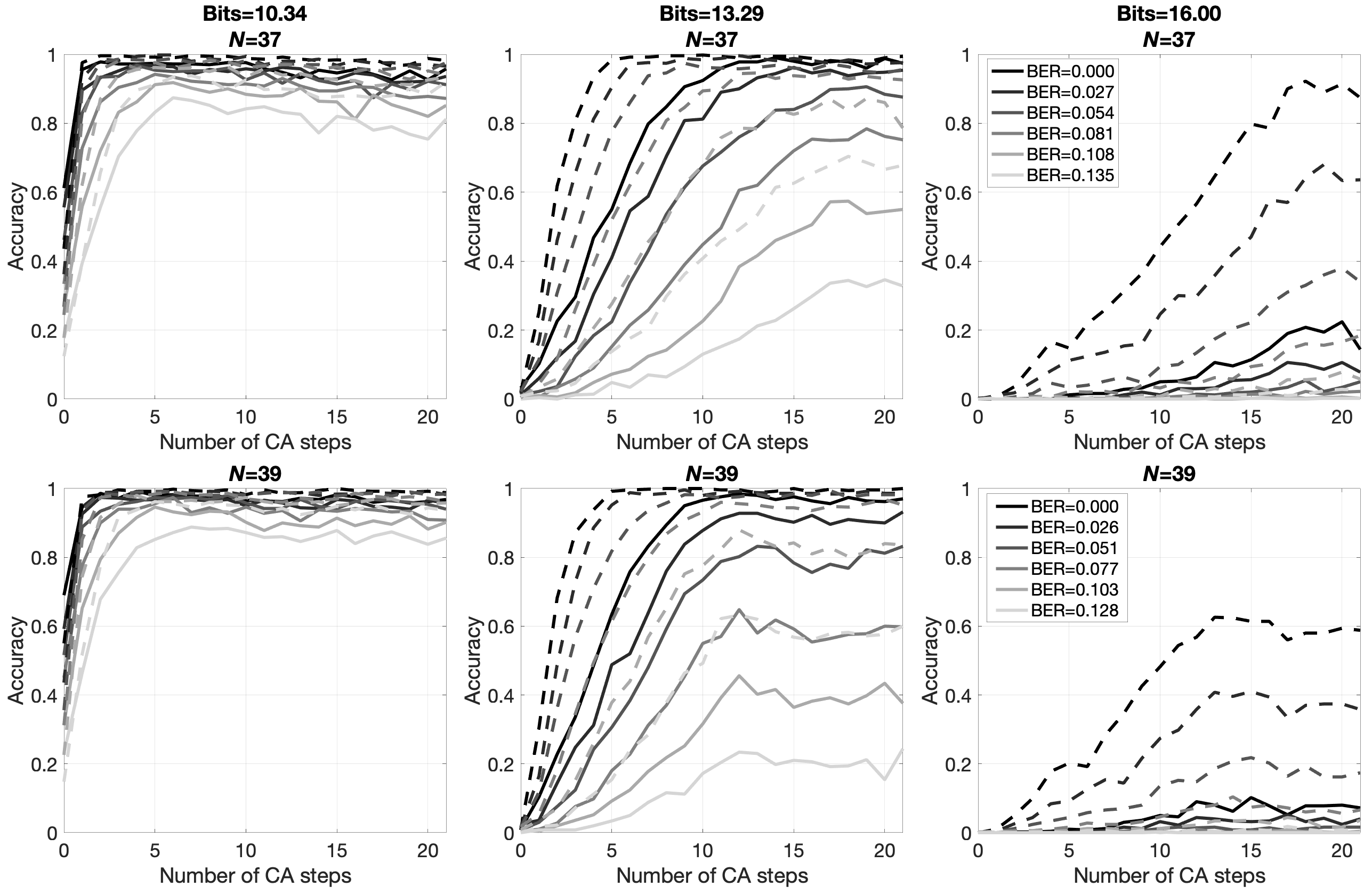}
\caption{
The usage of the CA90 expanded representations in the resonator network in the case when the initial short seed might have errors. 
The upper panels report the case for $N=37$ while the lower panels correspond to $N=39$.
The noise was introduced in the form of bit flips.
The number of bit flips was in the range $[0, 5]$ with step $1$.
The legends show the corresponding Bit Error Rates relative to $N$. 
The solid lines depict the resonator network with $4$ factors while the dashed lines depict the resonator network with $3$ factors.
Columns correspond to different amount of information carried by a vector, which was determined by the size of an item memory for one factor. 
The sizes of item memories for resonator networks with three and four factors were set to approximately match each other in terms of amount of information.
The reported values were averaged over $100$ simulations randomizing initial short seeds as well as introduced bit flips.
}
\label{fig:res:errors}
\end{figure*}

%\phantom{text} 
%\newpage
\subsection{Resonator network factoring in the case of errors}

%\todo[inline, color=yellow]{
%Report behavior of resonator for errors
%}

In order to examine the capabilities of CA90 expanded representations when the initial short seeds were subject to errors, we performed simulations for two dimensionalities of initial short seeds ($N=37$ and $N=39$).
Similar to the experiments in Fig.~\ref{fig:res:error:less}, we used the resonator network to reconstruct a randomly chosen combination of factors.   
The difference was that some bit flips were added to the initial short seed (i.e., vector to be factored) where the number of bit flips was in the range $[0, 5]$ with step $1$. 
The results are reported in Fig.~\ref{fig:res:errors}.
To minimize the noise introduced by CA90 computations we only used steps (\textit{x}-axis in Fig.~\ref{fig:res:errors}) of the form $2^j$ (cf. Fig.~\ref{fig:ca:1error}) to expand the dimensionality. 
Columns in Fig.~\ref{fig:res:errors} correspond to different amounts of information carried by the vector to be factored. 

The experiments were done for two configurations of the resonator network: with $3$ factors (dashed lines) and with $4$ factors (solid lines).
Clearly, when given the same conditions, the resonator network with $3$ factors outperforms the one with $4$ factors. 
This observation is in line with the expected behavior of the resonator network. 
It should be noted, however, that the resonator network with $3$ factors requires larger item memories in order store the same amount of information. 
For example, for $16.00$ bits in the case of $4$ factors the size of individual item memory was $16$ while in the case of $3$ factors it was $40$, i.e., the resonator network with $3$ factors required about $2.5$ times more memory. 
Thus, the use of a resonator network with fewer factors results in a better performance but it requires more memory to be allocated.

We also see that even in the absence of errors (BER=0.00) the accuracy is not perfect when a vector carries a lot of information because we are limited by the capacity of the resonator network -- which does fail at factorization when the size of the factorization problem is too large.
For example, for $16.00$ bits none of the expanded dimensionalities reached the perfect accuracy as opposed to the other two cases. 
Naturally, the inclusion of errors hurts the accuracy, but the performance degradation is gradual.

When comparing the performance of the resonator networks for the expanded vectors using all $21$ CA90 steps we made a counter intuitive observation that the performance for $N=37$ is better despite shorter vectors and higher Bit Error Rates (BER).
Recall from Fig.~\ref{fig:ca:periods} that the chosen grid sizes have different randomization periods: $87,381$ and $4095$ for $N=37$ and $N=39$,  respectively.
The longer randomization period for $N=37$ means that the use of $N=37$ provides more randomness for large number of CA90 steps. This is the main reason for the counter intuitive observation that the use of shorter seed at higher BER resulted in a better performance. 
When taking into account only the steps of the form $2^j$ the corresponding randomization period results in about $16.41$ and $12.00$.
These are exactly the values for which the performance of the resonator networks starts to saturate since concatenating additional dimensions after the randomization period stops adding extra randomness.

%% file: Discussion.tex
%\newpage
\section{Discussion}
\label{sect:discussion}

\subsection{Summary of our results}

%Talk about class 3 automata and their randomizing behaviour

The use of CA computations for the generation of random numbers is not new, for instance, a  
seminal work~\cite{WolframRandom1986} has proposed to generate random sequences with CA. 
Numerous studies followed on building CA-based pseudo-random number generators, e.g.,~\cite{SantoroSearch2007}, see \cite{DascaluCARandomization2018} for a recent overview of this work, some of it specifically investigating CA90.
%so the interested readers are kindly referred to this article for a general overview. 
Here, we focused on how collective-state computing can benefit from the randomness produced by CA90.
%can form the dense representations required in these models. 
%Naturally, we do not claim here that CA90 should be seen as a mechanism acting as, e.g., a cryptographically secure pseudo-random number generator but the point is that on a large scale the produced patterns seems random enough for practical purposes in collective-state computing.
Our results are based on a key observation 
%An interesting methodological observation is 
that collective-state computing typically relies on high-dimensional random representations, which have to be initially generated and stored for, e.g., being accessible by nearest neighbor searches during compute time. In many models, the representations are dense binary random patterns. 
Rather than storing the full random representations in item memory, we proposed to store just short seed patterns, and to use CA90 for re-materializing the full random representations when required.
The usage of CA90 expanded high-dimensional representations was demonstrated in the context of RC and VSAs.  
Our results provided empirical evidence that the expansion of representations on-demand (re-materialization solution) is functionally equivalent to storing the full i.i.d. random representations in the item memory (memorization solution).

%{\color{red}
%... [DESCRIBE WHAT ARE THE CONCRETE OUTCOMES, WHAT INTEGERS TO USE VERSUS TO AVOID ETC]
%}

Specifically, we  
%In this article we have explored how an elementary cellular automaton with rule 90 (CA90) can be used for on-demand generation of random distributed representations for collective-state computing models allowing for the space-time tradeoff. 
have shown that the randomization period of CA90 is closely connected to its periodic cycle length and depends of the size of the grid. 
We provided the exact relation between the grid size and the length of the randomization period. 
The general trend is that larger grid sizes yield longer randomization periods.
However, period length depends on number-theoretic properties of the grid size integer. 
In general, odd numbered grid sizes have longer randomization periods than even numbered. 
In particular, one should avoid grid sizes that are powers of two ($2^j$), as they have the shortest randomization period. 
%Next, in terms of length of randomization period are even grid sizes followed by odd sizes. 
The longest periods are obtained when the size of the grid is a prime number.  
Thus, given a memory constraint, it is best to choose the largest prime within the constraint.

We have also demonstrated that it is possible to use the expansion even in the presence of errors in the short seed patterns.
Unfortunately, CA90 introduces additional errors to the ones present in the seed pattern so the error rate after CA90 increases. 
The distribution of introduced errors is, however, not uniform -- some of the steps introduce more errors than the others. 
We have shown that the lowest amount of errors (cf. Fig.~\ref{fig:ber:ca:cur}) is introduced for CA90 steps that are powers of two ($2^j$).
Thus, in order to minimize the errors in the expanded representation one should use only steps of the form $2^j$. 
This, of course, limits the possibilities for expansion as practically not that many steps of the form $2^j$ can be computed (e.g., we used up to $20$ in the experiments).

%\todo[inline, color=yellow]{
%Remark: Randomness for computation ``frozen'' randomness\\
%How it relates to the use of noise sources with the idea of sampling, i.e., parameters are generated from some probability distribution
%}

\subsection{Related work}

\subsubsection{Combining reservoir computing, vector symbolic architectures and cellular automata}

%The combination of CA~\cite{Wolfram} computation with RC~\cite{LukoseviciusRC2009} and VSAs~\cite{Kanerva09} has been proposed in earlier work~\cite{YilmazMachine2015, YilmazSymbolic2015}.
%, building on some earlier work exploring this combination%RC, and VSA. 
%Several works done at the intersection of CA 
%These concepts were first combined together and explored in~
It has been demonstrated recently~\cite{FradyTheory2018, KleykointESN2017} that 
echo state networks~\cite{LukoseviciusPracticalESN2012}, an instance of RC, 
can be formulated in the VSA formalism. 
CA have been first introduced to RC and VSA models for expanding low-dimensional feature representations into high-dimensional representations to improve classification~\cite{YilmazMachine2015, YilmazSymbolic2015}. 
Due to the local interactions in CA, the evolution of the initial state (i.e., low-dimensional representation) over several steps produces a richer and higher-dimensional representation of features, while preserving similarity.
This method was applied to activation patterns from neural networks~\cite{YilmazMachine2015}, and to manually extracted features~\cite{KarvonenFPGA_CA_HD2019}.  The expanded representations were able to improve classification results for natural~\cite{YilmazMachine2015} and medical~\cite{KleykoModality2017} images. 
Similar to these works, our approach also employs
CA to expand the dimension of representations. However, we have applied expansion not to feature vectors, but just to i.i.d. random seed patterns. All we need is the property of CA90 that the resulting high-dimensional vectors are still pseudo-orthogonal. 
In our study, similarity preservation is only required if the random seed patterns contain errors.

Our work is most directly inspired by~\cite{SchmuckHardwareOptimizations2019} and~\cite{KleykoBrainsCA2017} who both used CA to expand item memory with short~\cite{SchmuckHardwareOptimizations2019} or long~\cite{KleykoBrainsCA2017, McDonaldReplicationCA2019} i.i.d. random seed patterns.
In~\cite{SchmuckHardwareOptimizations2019} the expansion was done with the CA30 rule, which is known to exhibit chaotic behaviour. 
Here, as in~\cite{KleykoBrainsCA2017}, we used the CA90 rule. 
%As in~\cite{SchmuckHardwareOptimizations2019}, we focus on expanding random representations via CA computations but while~\cite{SchmuckHardwareOptimizations2019} 
%This is similar to the focus of this article with the exception that we
%In contrast to~
%studied CA30, we used CA90. Rule 
For collective-state computing, CA90 has the great advantage that it distributes over the binding and the cyclic shift operation. 
We have seen this advantage in action when studying the resonator network in Section~\ref{sec:res:resnet:el}.
Since CA90 distributes over the binding operation, it was possible to expand the collective-state (i.e., the input vector (\ref{eq:factors}) with factors) on-demand during the factorization. 
%In particular, this work contributes to new knowledge by  providing the rigorous relation between the randomization period and the size of the grid. 
%Work~\cite{KleykoBrainsCA2017} differs from the studies described so far in the way how the CA computations were used.
%In~\cite{KleykoBrainsCA2017, McDonaldReplicationCA2019}, the randomization property of CA90 was used to reproduce an item memory with high-dimensional vectors making a unique clone in such a way that the reproduced item memory would be pseudo-orthogonal to the original one. 
Going beyond~\cite{KleykoBrainsCA2017, McDonaldReplicationCA2019}, we 
%not only applied %the randomization property  of 
also systematically explored the randomization properties of CA90, such as 
the length of the randomization period.
%by linking it to the size of the grid. 
%{\color{red}
%[WHAT DO YOU WANT TO SAY ABOUT THIS]
%}
Moreover, none of the previous work has studied the randomization behaviour of CA90 in the presence of errors in the initial seed. 
%Finally, for several collective-state computing scenarios this article provides empirical evidence that CA90 expanded representations are functionally equivalent to the ones from pseudo-random number generators.

\subsubsection{Other computation methods that use randomness}

%Connection to hashing
A complementary approach of computing with randomness is sampling-based computation~\cite{orban2016neural}. This approach differs fundamentally from collective-state computing, which exploits a concentration of mass phenomenon of random patterns making them pseudo-orthogonal. 
Once generated, a fixed set of random patterns can serve as unique and well distinguishable identifiers for handling variables and values during compute time.
%randomness in memorized patterns, i.e., randomness is created in an initial step and then kept frozen during computation. 
In contrast, in the sampling-based computation each compute step produces independent randomness to provide good mixing properties. 
Good mixing ensures that even a small set of samples is representative for the entire probability distribution and, therefore, constitutes a compact, faithful representation (ch. 29 in~\cite{mackay2003information}).
%{\color{red}[ADD BASIC REFERENCE EXPLAINING THIS]}.  
%or noise sources with the idea of sampling, e.g., some parameters of a system being constantly sampled from a probability distribution. 
%Collective-state and sampling-based approaches of computation may
%these two different approaches are related? Should they 
%be considered as two different extremes how randomness can be used. One uses pseudo-randomness to create unique, static representations for elements in the computation. The other uses randomness to represent probability distributions by a set of independent samples. 
We should add that the ``frozen'' randomness in VSA can be used to form different types of compact representation of probability distributions.   
%There are, however, connections as there are recent works in VSA domain, which demonstrate how to represent a histogram or probability distribution in a distributed way.
For example, a combination of binding and bundling can constitute compact representations of large histograms describing the $n$-gram statistics of languages~\cite{JoshiNgrams2016}. 
The advantage of such a representation is that it is a vector of the same fixed dimension as the atomic random patterns, somewhat independent of the number of non-zero entries in the histogram.
%such embedding is that the dimensionality of the representation does not depend directly on the size of the histogram.
%In turn, it allows building more compact and hardware-friendly models compared to the localist representation, which was demonstrated for the problems of language identification~\cite{JoshiNgrams2016, PSI19, RahimiLPHD} and text classification~\cite{HyperEmbed, Rasti2016}.
%Yet unanswered question is whether and how the distributed representations of probability distributions could be used for sampling from them. 

\subsection{Future work}

\subsubsection{Potential for hardware implementation}

The space-time or memory-computation tradeoff introduced by the inclusion of CA 
%We demonstrate that CA90 allows the space-time tradeoff. This tradeoff 
can be used to optimize the implementation of collective-state computations on hardware. 
%but it remains to be investigated how it exactly affects computations when implemented in hardware. 
Of course, this optimization depends on  
%computation/memory tradeoff can only be investigated 
the context of a computational problem and a particular hardware platform, which is outside the scope of this article.
%In other words, what are exact quantitative characteristics of the tradeoff in terms of increasing energy expenditures and slowing down the overall system to perform CA90 computations?\\ This question, however, is implementation dependent.
The optimized hardware implementation of models we described involves another interesting topic for future research, the question how to parallelize the computation of CA90.
%Another interesting direction for future work is the details of circuits realizing the tradeoff in hardware. 
While the implementation of CA90 in FPGA is quite straightforward, see equation (\ref{eq:ca90}), the implementation with neural networks and on neuromorphic hardware~\cite{Loihi18} is still an open problem. 
%Being said that, it should be admitted that while this article does not provide answers to these questions without this work the questions above are not even relevant as only here it is shown that space-time tradeoff with CA90 is a viable way to go.

\subsubsection{Integration of CA computations in neural associative memories}
Another interesting future direction is to investigate how associative memories~\cite{SurveyAM17} can trade off synaptic memory with neural computation implementing the CA. 
Such CA-based approaches could be compared to other suggestions in the literature how to replace memory by computation, e.g.,~\cite{knoblauch2010zip}.

%\todo[inline, color=yellow]{
%Implementation dependent \\
%What does tradeoff mean in terms of computation?\\
%Are there interesting network implementations of this?\\
%What exactly is tradeoff in terms of slowing down computation?\\
%Without this work the questions above are not even relevant -> Is this a viable way to go? \\
%}

%\todo[inline, color=yellow]{
%Think of neural implementation.
%}

%\todo[inline, color=yellow]{
%Look for more interesting mode of clean-up memory
%}

%\todo[inline, color=yellow]{
%Check CA with complex numbers
%}

%% file: intESN.tex
\section{Integer echo state network}
\label{sect:intesn}

The integer echo state network (intESN) has been proposed in~\cite{KleykointESN2017}) as a light weight alternative to the conventional ESN~\cite{LukoseviciusPracticalESN2012}. 
For the sake of simplicity, we introduce it here using the memory buffer task from Section~\ref{sec:mem:buf}.
The memory buffer task~\cite{FradyTheory2018,PlateBook} has two stages: memorization and recall.
At the memorization stage, at every timestep $m$  the intESN  stores a symbol $\textbf{s}(m)$ from the sequence of symbols $\textbf{s}$ to be memorized.
The number of unique symbols (i.e., alphabet size) is denoted as $D$.
The symbols are represented using $K$-dimensional random bipolar dense vectors stored in the item memory $\mathbf{H}  \in \{-1,1\}^{K \times D}$.
Thus, at every timestep $m$ the intESN is presented with the corresponding $K$-dimensional vector $\mathbf{H}_{\textbf{s}(m)}$, which is added to the hidden layer (i.e., the reservoir storing the collective state) of the intESN ($\textbf{x} \in \mathbb{Z}^{K \times 1}$). 
The state of the hidden layer at timestep $m$ (denoted as $\textbf{x}(m)$) is updated as follows: 
\noindent
\begin{equation}
\textbf{x}(m)= f_\kappa ( \rho(\textbf{x}(m-1)) +  \mathbf{H}_{\textbf{s}(m)}  ),
\label{eq:intESN}
\end{equation}
\noindent
where $\textbf{x}(m-1)$ is the previous state of the hidden layer at timestep $m-1$;
$\rho$ denotes the permutation operation (e.g., cyclic shift to the right), which in the intESN acts as a simple variant of a recurrent connection matrix;
$f_\kappa (*)$ is a clipping function defined as:
\noindent
\begin{equation}
f_\kappa (x) = 
\begin{cases}
-\kappa & x \leq -\kappa \\
x & -\kappa < x < \kappa, \\
\kappa & x \geq \kappa
\end{cases}
\label{eq:clipping}
\end{equation}
\noindent
where $\kappa$ is a configurable threshold parameter. 
In the scope of the intESN, the clipping function acts as a nonlinear activation function, which keeps the values of the hidden layer in the limited range determined by $\kappa$.
In practice, the value of $\kappa$  regulates the recency effect of the intESN.

At the recall stage, the intESN uses the current collective state in its hidden layer $\textbf{x}(m)$ as the query vector to retrieve the symbol stored $d$ steps ago, where $d$ denotes delay. 
The recall is done by using the readout matrix for particular $d$, which contains one $K$-dimensional vector per each symbol.
The readout matrix is denoted as $\textbf{W}^{d} \in \mathbb{R}^{D \times N}$ and the recall is done as: 
\noindent
\begin{equation}
\hat{\textbf{s}}(m-d)=\argmax ( \textbf{W}^{d} \textbf{x}(m)  ),
\label{eq:recall}
\end{equation}
\noindent
where $\argmax(\cdot)$ returns the symbol with the highest postsynaptic sum among the output neurons for the chosen delay $\textbf{W}^{d}$ value and the given collective state $\textbf{x}(m)$.

There are several approaches to form the readout matrix. 
In the experiments, we used the approach, which is most common in ESN that is to obtain $\textbf{W}^{d}$ via solving a linear regression on a given training sequence and the corresponding states of the hidden layer. 

%% file: Resonator.tex
\section{Resonator network}
\label{sect:resnet}

Recall that in Section~\ref{sect:vsas} when introducing the binding operation (\ref{eq:bind}) only a pair of vectors was bound: $\textbf{x}  \oplus \textbf{y}$.
There are, however, practical cases when several vectors should be bound, e.g., $\textbf{z} = \textbf{v}  \oplus \textbf{w}  \oplus\textbf{x}  \oplus \textbf{y}$.
A concrete example when this design is used is a representation of $n$-grams by vectors~\cite{JoshiNgrams2016}.
Here we assume that each component (factor; denoted as $\mathbf{f}_i$) comes from a separate item memory ($\mathbf{H}_1, \mathbf{H}_2, ...$), which is called factor item memory, i.e., a general example of a vector with four factors is 
$\mathbf{f}_1  \oplus \mathbf{f}_2  \oplus \mathbf{f}_3  \oplus \mathbf{f}_4$.
Note that the task of factoring one of the components from a vector representing the result of binding of two vectors is a relatively simple task as, e.g, $\textbf{x}  \oplus \textbf{y} \oplus \textbf{x} = \textbf{y}$.
This task becomes much more complex (complexity grows exponentially) in the case when several vectors are bound together.
However, there is a very recent work~\cite{KentResonatorNetworks2019, ResPart1}, which proposed an elegant mechanism called the resonator network to address the challenge of the factoring. 
In the nutshell, the resonator network~\cite{ResPart1} is a novel recurrent neural network design that uses VSAs principles to solve combinatorial optimization problems. 
To factor the components from a vector $\textbf{z}$ representing the bindings of several vectors, the resonator network uses several populations of units,  $\mathbf{\hat{f}}_1(t), \mathbf{\hat{f}}_2(t), ...$, each of which tries to infer a particular factor from the input vector. 
Each population, called a resonator, communicates with the input vector and all other neighboring populations to invert $\textbf{z}$. 

In essence, each resonator can hold multiple weighted guesses for a vector from each factor item memory through the VSAs principle of superposition, which is used for the bundling operation. 
Each resonator then also receives guesses for factors from other resonators. 
These guesses from the other resonators are used to invert the input vector and infer the factor of interest in the given resonator.
The principle of superposition allows a population to hold multiple estimates of factor identity and test them simultaneously. 
The cost of superposition is a crosstalk noise. 
The inference step is, thus, noisy when many guesses are tested at once. 
But the next step is to use the item memory $\mathbf{H}_i$  to remove the extraneous guesses that do not fit. 
Thus, the guess $\mathbf{\hat{f}}_i$ for each factor is cleaned-up by constraining the resonator activity only to the allowed atomic vectors stored in $\mathbf{H}_i$.
The main dynamics of the resonator network are best described by the iterative update process. The resonator dynamics can be solved by following the VSAs operations.  
Through VSAs operations, the representation of an input vector $\textbf{z}$ can be approximately inverted.  
Thus, to factorize $\textbf{z}$, we can construct an equation for each factor as follows:
\noindent
\begin{equation}
\begin{split}
&\mathbf{\hat{f}}_1(t+1)= f \Big( \mathbf{H}_1 \mathbf{H}_1^T (\textbf{z} \oplus  \mathbf{\hat{f}}_2(t)  \oplus  \mathbf{\hat{f}}_3(t) \oplus  \mathbf{\hat{f}}_4(t) ) \Big) \\
&\mathbf{\hat{f}}_2(t+1)= f \Big( \mathbf{H}_2 \mathbf{H}_2^T (\textbf{z} \oplus  \mathbf{\hat{f}}_1(t)  \oplus  \mathbf{\hat{f}}_3(t) \oplus  \mathbf{\hat{f}}_4(t) ) \Big) \\
&\mathbf{\hat{f}}_3(t+1)= f \Big( \mathbf{H}_3 \mathbf{H}_3^T (\textbf{z} \oplus  \mathbf{\hat{f}}_1(t)  \oplus  \mathbf{\hat{f}}_2(t) \oplus  \mathbf{\hat{f}}_4(t) ) \Big) \\
&\mathbf{\hat{f}}_4(t+1)= f \Big( \mathbf{H}_4 \mathbf{H}_4^T (\textbf{z} \oplus  \mathbf{\hat{f}}_1(t)  \oplus  \mathbf{\hat{f}}_2(t) \oplus  \mathbf{\hat{f}}_3(t) ) \Big) \\
\end{split}
\label{eqn:resnet}
\end{equation}
\noindent
Note that according to (\ref{eqn:resnet}) to infer, e.g., the second factor, we also need to know the identities of the first, third, and fourth factors, which are not known in advance. 
Therefore, the resonator network relies on an iterative approach where the guesses for all factors are being inferred simultaneously. 
Thus, the vectors $\mathbf{\hat{f}}_1$, $\mathbf{\hat{f}}_3$ and $\mathbf{\hat{f}}_4$, can be thought to hold multiple guesses for their factors in superposition. Only one of these guesses will be correct, however. 
Thus, the interaction of the incorrect guesses with the inference procedure will result in more terms that act as the crosstalk noise. 
VSAs are often faced with these noisy inference procedures, and they use the item memory to remove the noise. 
This is simply a comparison between the inference of $\mathbf{\hat{f}}_2$ with all the possible atomic vectors, which are stored in the item memory $\mathbf{H}_2$ for this factor. 
Another way of understanding this clean-up step is that the resonator network is constrained to the span of the vectors stored in the item memory. 
Finally, a regularization step (denoted as $f(*)$) is needed, which can be either normalization or bipolarization.
Successive iterations of this inference  and clean-up procedure (\ref{eqn:resnet}), eliminate the noise as the factors become identified and find their place in the input vector. 
When the factors are fully identified, the resonator network reaches a stable equilibrium and the factors can be read out from the stable activity pattern.

When running the resonator network, at first, the system appears to bounce around stochastically, with the inferences being dominated by the crosstalk noise. Eventually (in a regime within the ``operational capacity''~\cite{ResPart1}), the system has a moment of insight and rapidly hones in on the solution and the dynamics converges to a stable equilibrium. 
Once it converges, the factor corresponding to the atomic vector with the largest similarity to the final state of each estimator is taken as the output.